\documentclass[lettersize,journal]{IEEEtran}
\usepackage{amssymb}
\usepackage{amsthm}
\usepackage{amsmath}
\usepackage{mathrsfs}
\usepackage{stmaryrd}
\usepackage{graphicx,amsfonts,listings,xcolor,colortbl}
\usepackage{multirow}
\usepackage{booktabs}
\usepackage{tabularx}
\usepackage{blkarray}
\usepackage{arydshln}
\usepackage{subfigure}
\usepackage{fancyhdr}
\usepackage{setspace}
\usepackage{multirow}
\usepackage{color}
\usepackage{xcolor}
\usepackage{threeparttable}

\SetSymbolFont{stmry}{bold}{U}{stmry}{b}{n}

\usepackage{multicol}
\usepackage{times}
\usepackage{soul}
\usepackage{refcount}
\usepackage{url}
\usepackage[utf8]{inputenc}
\usepackage{makecell}
\usepackage{float}
\usepackage{stfloats}
\usepackage[OT1]{fontenc}

\usepackage{stmaryrd} \SetSymbolFont{stmry}{bold}{U}{stmry}{m}{n} 
\usepackage{bm}
\usepackage{algorithm}
\usepackage{algpseudocode}
\usepackage{diagbox}
\usepackage[justification=centering]{caption}

\renewcommand{\arraystretch}{1.3}

\begin{document}

\title{Kernel Alignment-based Multi-view Unsupervised Feature Selection with Sample-level Adaptive Graph Learning}

\author{Yalan~Tan,~Yanyong~Huang*,~Zongxin~Shen,~Dongjie Wang,~Fengmao~Lv,~and Tianrui Li,~\IEEEmembership{Senior Member,~IEEE}
    \thanks{This work has been submitted to the IEEE for possible publication. Copyright may be transferred without notice, after which this version may no longer be accessible.}
    \thanks{Yalan Tan, Yanyong Huang, and Zongxin Shen are with the Joint Laboratory of Data Science and Business Intelligence, School of Statistics and Data Science, Southwestern University of Finance and Economics, Chengdu 611130, China (e-mail: tyl2019@smail.swufe.edu.cn; huangyy@swufe.edu.cn; zxshen@smail.swufe.edu.cn), Yanyong Huang is the corresponding author;}
    \thanks{Dongjie Wang is with the Department of Electrical Engineering and Computer Science, University of Kansas, Lawrence, KS 66045, USA (e-mail: wangdongjie@ku.edu);}
    \thanks{Fengmao~Lv  and Tianrui Li are with the School of Computing and Artificial Intelligence, Southwest Jiaotong University, Chengdu 611756, China (e-mail: fengmaolv@126.com, trli@swjtu.edu.cn).}
}

\markboth{Journal of \LaTeX\ Class Files,~Vol.~14, No.~8, August~2021}%
{Shell \MakeLowercase{\textit{et al.}}: A Sample Article Using IEEEtran.cls for IEEE Journals}

\maketitle

\begin{abstract}
Although multi-view unsupervised feature selection (MUFS) has demonstrated success in dimensionality reduction for unlabeled multi-view data, most existing methods reduce feature redundancy by focusing on linear correlations among features but often overlook complex nonlinear dependencies. This limits the effectiveness of feature selection. In addition, existing methods fuse similarity graphs from multiple views by employing sample-invariant weights to preserve local structure. However, this process fails to account for differences in local neighborhood clarity among samples within each view, thereby hindering accurate characterization of the intrinsic local structure of the data. In this paper, we propose a Kernel Alignment-based multi-view unsupervised FeatUre selection with Sample-level adaptive graph lEarning method (KAFUSE) to address these issues. Specifically, we first employ kernel alignment with an orthogonal constraint to reduce feature redundancy in both linear and nonlinear relationships. Then, a cross-view consistent similarity graph is learned by applying sample-level fusion to each slice of a tensor formed by stacking similarity graphs from different views, which automatically adjusts the view weights for each sample during fusion. These two steps are integrated into a unified model for feature selection, enabling mutual enhancement between them. Extensive experiments on real multi-view datasets demonstrate the superiority of KAFUSE over state-of-the-art methods.
\end{abstract}

\begin{IEEEkeywords}
Multi-view unsupervised feature selection, Kernel alignment, Cross-view consistency, Adaptive graph learning.
\end{IEEEkeywords}

\section{Introduction}
\IEEEPARstart{M}{ulti-view} data describe the same sample through heterogeneous features from multiple perspectives and are commonly encountered in real-world applications~\cite{komeili2020multiview,xiang2025class}. For example, in an image classification task, an image can be characterized by different visual descriptors, such as color similarity, RGB and HSV color histograms. Multi-view data are typically high-dimensional due to the integration of multiple distinct views, which can lead to the curse of dimensionality and degrade the performance of downstream tasks~\cite{C2IMUFS,sun2021joint}. In addition, acquiring labels is costly and labor-intensive, which often results in the presence of unlabeled data~\cite{shao2016online,tang2019cross}. Consequently, the effective reduction of dimensionality in multi-view unlabeled data to improve downstream task performance has become an urgent issue that needs to be addressed in many practical applications.

Multi-view unsupervised feature selection (MUFS)~\cite{zhang2019feature,liu2025latent} effectively addresses this problem by selecting a compact subset of representative features from multi-view unlabeled data. In recent years, various MUFS methods have been developed, which can be broadly categorized into two groups.  In the first group, features from different views are concatenated into a single view and then processed using conventional single-view-based feature selection approaches. Representative methods in this category include General Adaptive Feature Selection with Auto-weighting (GAWFS)~\cite{2025GAWFS}, and Robust Neighborhood Embedding (RNE) for unsupervised feature selection~\cite{liu2020RNE}. Although these methods can enhance feature selection performance to a certain extent, they treat each view independently and fail to exploit the intrinsic relationships among multiple views. Instead of stacking  features from different views, the second category of MUFS methods builds models directly from multi-view data to perform feature selection. Representative methods in this category include Unsupervised Kernel-based Multi-view Feature Selection (UKMFS)~\cite{2025UKMFS}. UKMFS constructs low-rank consistent graphs through self-representation and subsequently utilizes an unsupervised hashing-based model to obtain reliable pseudo binary labels for feature selection. Besides, Wang et al. have introduced a Weighted Low-rank Tensor Learning-based method (WLTL) for unsupervised multi-view feature selection, wherein clustering indicator matrices are stacked into a tensor, and a weighted tensor nuclear norm constraint is employed to derive consistent pseudo labels for guiding discriminative feature selection~\cite{2025WLTL}. In addition, Zhou et al. have developed a Consistency-Exclusivity guided Unsupervised Multi-view Feature Selection method, called CE-UMFS. This method employs matrix factorization to simultaneously learn a feature selection matrix along with both view-specific and unified coefficient matrices, while also incorporating the Hilbert-Schmidt independence criterion to maintain exclusivity among views~\cite{CE2024consistency}.

\begin{figure*}[t]
	\centering
	\includegraphics[width=0.9\textwidth, keepaspectratio]{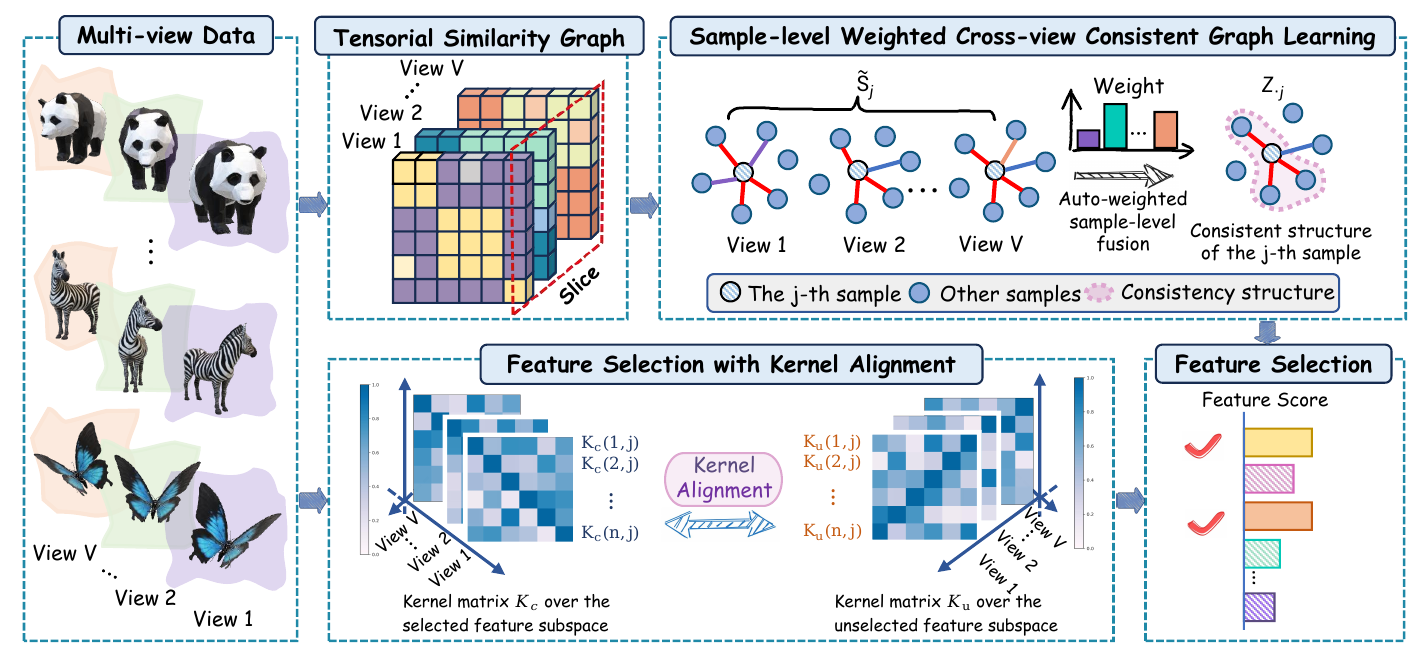}
	\caption{The framework of the proposed KAFUSE.}
	\label{Fig1}
\end{figure*}

Although these methods focus on enhancing the discriminability of selected features, they do not sufficiently consider redundancy, which may result in a compact subset of selected features that remains correlated and redundant, thus diminishing the overall effectiveness of feature selection. There have been attempts to mitigate feature redundancy in existing methods. For example, CDMvFS enforces orthogonality on feature-weighted data to reduce redundancy in the original dataset~\cite{CDMvFS2024multi}. Besides, MAMFS imposes orthogonality on projected data to eliminate feature redundancy~\cite{MAMFS2021multilevel}. However, these methods address feature redundancy by focusing solely on linear correlations between features, while neglecting the complex nonlinear dependencies present in real-world data. This ultimately limits their ability to address nonlinear redundancy. 

Furthermore, previous studies have demonstrated that leveraging cross-view information to construct a consistent similarity graph is beneficial for improving the performance of feature selection~\cite{wan2020adaptive}. Nevertheless, existing methods use sample-invariant weights to fuse similarity graphs from different views to obtain a consistent graph~\cite{fang2023joint,ke2023clustering}, while overlooking that the clarity of local neighborhood structures may differ among samples within each view. For example, in image clustering tasks, color and texture can be considered as two distinct views. In the color view, certain animal samples such as black panthers and polar bears can be easily distinguished based on their color information, which leads to well-separated clusters with clear local neighborhood structures. In contrast, animals like zebras, which have mixed black-and-white patterns, are more difficult to differentiate in the color view and tend to exhibit ambiguous neighborhood structures. As a result, assigning identical fusion weights to all samples, as traditional methods do,  may cause clusters that should be clearly separated to become blurred, thereby making neighborhood relationships less reliable. This limits the accurate capture of the intrinsic local geometric structure of the data and ultimately degrades feature selection performance.

To address the above issues, we propose a novel MUFS method, referred to as Kernel Alignment-based multi-view unsupervised FeatUre selection with Sample-level adaptive graph lEarning (KAFUSE).  Specifically, we first use kernel alignment with an orthogonal constraint to reduce feature redundancy in both linear and nonlinear relationships, enabling the identification of discriminative and uncorrelated features. Subsequently, the similarity graphs from different views are concatenated into a tensor. We then learn a cross-view consistent similarity graph by applying sample-level fusion to each slice of the tensor, which automatically adjusts the view weights for each sample during the fusion process. The learning of both the cross-view consistent similarity graph and the view-specific similarity graphs is integrated into the feature selection process, allowing them to enhance each other. Furthermore, we develop an alternating iterative optimization algorithm to solve the proposed model and demonstrate the effectiveness of our method through comprehensive comparisons with state-of-the-art approaches on multiple real-world multi-view datasets. The overall framework of the proposed KAFUSE method is illustrated in Fig.~\ref{Fig1}.

The major contributions of this paper are summarized as follows:
(\romannumeral1) Instead of assigning the same fusion weight to all samples within each view, we design  a cross-view consistent similarity graph learning model with a sample-level view weighting mechanism that automatically adjusts the view weights for each individual sample, allowing for better preservation of the data's local structure. (\romannumeral2) Kernel alignment with an orthogonal constraint is employed to mitigate feature redundancy in both linear and nonlinear relationships, thereby facilitating the selection of discriminative and uncorrelated features. (\romannumeral3) An effective alternating optimization algorithm is developed, and comprehensive experimental comparisons demonstrate the superiority of the proposed method over competing methods.

The remainder of this paper is organized as follows. Section II presents a brief overview of related work on MUFS. In Section III, we provide a detailed introduction to the proposed KAFUSE method and present an effective iterative optimization algorithm to solve the proposed problem, along with convergence and complexity analyses. In Section IV, we conduct a series of experiments to demonstrate the effectiveness of the proposed method. Section V concludes the paper.

\section{Related Works}
In this section, we provide a brief review of representative works on MUFS, which are categorized into single-view-based and multi-view-based methods. Single-view-based unsupervised feature selection methods typically concatenate features derived from different views into a single dataset, upon which conventional unsupervised feature selection algorithms are subsequently applied. Among the representative methods in this category, GAWFS uses nonnegative matrix factorization to generate a nonnegative indicator matrix and combines this with adaptive graph learning to construct a similarity graph, which then guides the learning of a feature weighting matrix for feature selection~\cite{2025GAWFS}. RNE reconstructs each data point from its local neighbors and minimizes the reconstruction error via $\ell_{1}$-norm regularization, while performing feature selection through a feature-indicator matrix~\cite{liu2020RNE}. To improve robustness against outliers, GLoRSS measures reconstruction error with correntropy and achieves feature selection in a learned sparse subspace~\cite{zhou2017maximum}. When dealing with multi-view data, single-view-based unsupervised feature selection methods often combine features from different views without considering their interrelationships, which can degrade the effectiveness of feature selection.

In contrast to the view concatenation strategy employed in single-view-based methods, multi-view unsupervised feature selection methods directly process multi-view data and leverage underlying inter-view correlations to improve performance. Among these methods, CDMvFS stands out by generating mutually exclusive graphs to strengthen complementarity across views and integrating graph learning with consensus clustering to enforce indicator matrix consistency~\cite{CDMvFS2024multi}. MFSGL adaptively learns view weights to build a consensus similarity graph across multiple views. By imposing a rank constraint, it optimizes the similarity matrix to facilitate more effective feature selection~\cite{MFSGL2021autoweighted}. To capture both within-view and cross-view information, MAMFS incorporates weighted projections for each view along with a shared projection, and subsequently learns a collaborative graph that adapts to $k$-nearest neighbors within subspaces to facilitate feature selection~\cite{MAMFS2021multilevel}. CFSMO constructs view-specific initial similarity matrices using random walks to preserve higher-order neighborhood information, and projects multi-view data into a shared latent representation to enable effective feature selection~\cite{CFSMO2024multi}. To preserve the intrinsic clustering structure, UKMFS learns a multi-view consistent graph representation by adaptively estimating view weights. It also exploits representative features from each view by incorporating view-specific projection matrices and weakly-supervised labels obtained from multi-view binary hashing codes~\cite{2025UKMFS}. CE-UMFS applies the Hilbert-Schmidt Independence Criterion to view-specific coefficient matrices and enforces a low-rank constraint on the unified coefficient matrix to identify both consistent and exclusive features~\cite{CE2024consistency}. WLTL selects features by embedding clustering indicator matrices into a tensor representation and applying weighted tensor nuclear norm regularization to capture high-order correlations across different views. 

Although existing methods have shown promising results in practical applications, they still face two major limitations. First, their ability to address feature redundancy is restricted, as they mainly focus on linear relationships among features and therefore overlook redundancy caused by nonlinear interactions. Second, when constructing a consistent similarity graph, they assign uniform weights to samples without accounting for differences in the clarity of local neighborhoods. This leads to a fused graph with blurred local neighborhood structures, ultimately resulting in suboptimal feature selection performance.

\section{The Proposed Method}
In this section, we begin by introducing the notations and definitions used throughout the paper. Next, we describe the proposed KAFUSE and the alternative iterative algorithm designed to solve the optimization problem. Finally, we analyze the algorithm's convergence and computational complexity.

\subsection{Notations and Definitions}
Throughout this paper, matrices are denoted by bold uppercase letters, vectors by bold lowercase letters, and scalars by regular lowercase letters. For a matrix $\bm{A} \in \mathbb{R}^{m \times n}$, $\bm{A}_{i \cdot}$, $\bm{A}_{\cdot j}$, and $\bm{A}_{i j}$ represent the $i$-th row, $j$-th column and $(i,j)$-th entry of $\bm{A}$, respectively. The trace of a matrix $\bm{A}$ is denoted by $\operatorname{Tr}(\bm{A})$, and its transpose by $\bm{A}^T$. The Frobenius norm of $\bm{A}$ is defined as $\|\bm{A}\|_F=\sqrt{\sum_{i=1}^{m} \sum_{j=1}^{n} \bm{A}_{ij}^{2}}$. Additionally, the vector $\bm{1}_n \in \mathbb{R}^{n \times 1}$ refers to an all-ones column vector, and the matrix $\bm{I}_c \in \mathbb{R}^{c \times c}$ denotes the identity matrix.

Given a multi-view dataset $\mathcal{X}=\{ \bm{X}^{(v)} \in \mathbb{R}^{d_v \times n}\}_{v=1}^{V}$ containing  $V$ views, where $\bm{X}^{(v)}$ is the data matrix of the $v$-th view with $d_v$ features and $n$ samples. Our goal is to identify the most  informative $l$ features from $\mathcal{X}$ in an unsupervised manner.

\subsection{Kernel Alignment-based Redundancy Reduction for MUFS}
Orthogonal least squares regression is a widely used machine learning technique that has been applied to various tasks, such as regression and clustering~\cite{souza2013regression,xu2021general}. In this study, we employ orthogonal least squares regression to select informative features from multi-view data, as described below:
\begin{equation}\label{3.2_2}
	\begin{aligned}
	\min_{\substack{~~\bm{W}^{(v)},\bm{b}^{(v)},\\\!\!\bm{\Lambda}^{(v)},\bm{F}}}\sum_{v=1}^{V}\| {\bm{W}^{(v)}}^T \bm{\Lambda}^{(v)} \bm{X}^{(v)} + &\bm{b}^{(v)} \bm{1}^T_n\! - \!\bm{F} \|_F^2\\
	\text{ s.t. }{\bm{W}^{(v)}}^T\!\bm{W}^{(v)}=\bm{I}_c,\bm{F}\bm{F}^T =&\bm{I}_c, \bm{\lambda}^{(v)} \in \{0,1\},
    \end{aligned}
\end{equation}
where $\bm{W}^{(v)}\!\!\in \mathbb{R}^{d_v \times c}$ denotes the projection matrix of the $v$-th view, $\bm{b}^{(v)} \in \mathbb{R}^{c \times 1}$ is the bias vector, and $\bm{F} \in \mathbb{R}^{c \times n}$ represents the clustering indicator matrix. In addition, $\bm{\lambda}^{(v)} \in \mathbb{R}^{d_v \times 1}$ is the feature selection indicator vector of the $v$-th view, where ${\lambda}^{(v)}_i = 1$ if the $i$-th feature is selected and ${\lambda}^{(v)}_i = 0$ otherwise. The diagnal matrix $\bm{\Lambda}^{(v)} \!\!=\! \operatorname{diag}(\bm{\lambda}^{(v)}) \in \mathbb{R}^{d_v \times d_v}$ is constructed from $\bm{\lambda}^{(v)}$. The orthogonal least squares regression model in Eq.~(\ref{3.2_2}) imposes an orthogonal constraint on the view-specific projection matrix \(\bm{W}^{(v)}\) to reduce linear redundancy among features. However, this orthogonal constraint can only guarantee linear independence and do not address nonlinear relationships between features, which limits the model's effectiveness in handling redundancy caused by nonlinear dependencies~\cite{wei2016nonlinear}.

To address the aforementioned issue, we use kernel matrices~\cite{liu2019multiple, cortes2012algorithms} to implicitly capture the latent nonlinear relationships within feature subspaces. Specifically, based on the feature selection indicator vector \(\bm{\lambda}^{(v)}\), we denote the data matrix containing the selected features as \(\bm{X}^{(v)}_c = \bm{\Lambda}^{(v)}\bm{X}^{(v)}\), while the data matrix containing the unselected features is denoted as \(\bm{X}^{(v)}_{u} = (\bm{I}-\bm{\Lambda}^{(v)})\bm{X}^{(v)}\). Then, we use a Gaussian kernel to measure the similarity between samples in the reproducing kernel Hilbert space (RKHS)~\cite{aronszajn1950theory}. The kernel matrices for the selected and unselected feature subspaces are calculated as follows, respectively: 
\begin{align}\label{3.2_3}
    &\left[\!\bm{K}^{(v)}_c\!\right]_{ij}\!\!\!\!=\operatorname{exp}(-\frac{1}{\sigma^2}\|\bm{\Lambda}^{(v)}\bm{X}^{(v)}_{\cdot i}-\bm{\Lambda}^{(v)}\bm{X}^{(v)}_{\cdot j}\|^2_2),\\
	&\left[\!\bm{K}^{(v)}_{u}\!\right]_{ij}\!\!\!\!=\operatorname{exp}(\!-\!\frac{1}{\sigma^2}\|(\bm{I}\!\!-\!\bm{\Lambda}^{(v)}\!)\bm{X}^{(v)}_{\cdot i}\!-\!(\bm{I}\!-\!\bm{\Lambda}^{(v)}\!)\bm{X}^{(v)}_{\cdot j}\|^2_2),
\end{align}
where $\sigma$ is the bandwidth parameter. After centering, these two kernel matrices can be represented as \(\bm{H}\bm{K}^{(v)}_c\bm{H}\) and \(\bm{H}\bm{K}^{(v)}_{u}\bm{H}\), where \(\bm{H} = \bm{I}_n - \frac{1}{n}\bm{1}_n\bm{1}_n^T\) is the centering matrix. Kernel alignment~\cite{cristianini2001kernel, wang2015overview} is then used to measure the alignment between the sample similarity structures induced by the nonlinear kernels constructed from the selected and unselected feature subspaces, as shown below:
\begin{equation}\label{3.2_5}
\max_{\bm{\Lambda}^{(v)}}\operatorname{Tr}(\bm{H}\!\bm{K}^{(v)}_c \!\bm{H}\!\bm{H}\!\bm{K}^{(v)}_{u}\!\bm{H})\!= \!\max_{\bm{\Lambda}^{(v)}}\operatorname{Tr}(\bm{H}\!\bm{K}^{(v)}_c\!\bm{H}\!\bm{K}^{(v)}_{u}).
\end{equation} 

A high degree of alignment between the centered kernel matrices of the selected features \(\bm{K}^{(v)}_c\) and the unselected features \(\bm{K}^{(v)}_{u}\) indicates that these two feature subsets produce highly similar data structures in their respective RKHS. In other words, the information captured by the selected features closely resembles that contained in the unselected feature set. Thus, by maximizing the alignment between the kernel matrices of the selected and unselected features, Eq.~(\ref{3.2_5}) facilitates the partitioning of highly redundant features that provide similar information into \(\bm{X}^{(v)}_c\) and \(\bm{X}^{(v)}_{u}\), thereby effectively reducing nonlinear redundancy within the selected feature subset. By integrating Eq.~(\ref{3.2_2}) with Eq.~(\ref{3.2_5}), We obtain the kernel alignment-based redundancy reduction module for MUFS as follows:
\begin{equation}\label{3.2_4}
	\begin{aligned}
	\min_{\bm{\Omega}_1}&\!\!\sum_{v=1}^{V}\!\bm{\theta}_v^2\| {\bm{W}^{(v)}}^T\!\!\!\! \bm{\Lambda}^{(v)}\!\! \bm{X}^{(v)} \!\!\!+ \!\bm{b}^{(v)}\!\bm{1}^T_n\!\!\!- \!\!\bm{F} \|_F^2\!\!-\!\bm{\omega}_v^r\!\operatorname{Tr}(\!\bm{H}\!\bm{K}_c^{(v)} \!\!\bm{H} \!\bm{K}_{u}^{(v)}\!)\\
	\text{s.t.}&~~{\bm{W}^{(v)}}^T\!\bm{W}^{(v)}=\bm{I}_c,\bm{F}\bm{F}^T =\bm{I}_c, \bm{\lambda}^{(v)} \in \{0,1\}.
    \end{aligned}
\end{equation}
where $\bm{\Omega}_1=\{\bm{W}^{(v)},\bm{\Lambda}^{(v)},\bm{b}^{(v)},\bm{F}, \bm{\theta}_v, \bm{\omega}_v\}$, $\bm{\theta}_v$ and $\bm{\omega}_v$ are the view weightes, and $r$ is the trade-off parameter. In Eq.~(\ref{3.2_4}), we simultaneously apply kernel alignment and an orthogonal constraint to capture both linear and nonlinear correlations, thereby promoting the selection of discriminative and uncorrelated features.

\subsection{Cross-view Consistent Similarity Graph Learning via Auto-weighted Sample-level Fusion}
Prior studies have demonstrated that the exploitation of cross-view consistency information to construct a consistent similarity graph between views is beneficial for enhancing the performance of MUFS~\cite{hou2017multi,tang2023unsupervised}. Most traditional methods construct a consistent similarity graph by employing sample-invariant weights to fuse similarity graphs from different views, thereby assigning identical fusion weights to all samples. However, within a given view, samples can vary in the clarity of their local neighborhood structures. Some samples have clearly defined neighborhoods, while others have ambiguous neighborhood relationships. As a result, assigning the same weights to all samples during graph fusion may result in a fused similarity graph with unreliable local neighborhood structures, thereby making it difficult to accurately preserve the intrinsic local structure of the data.

To address this limitation, we propose a novel method for cross-view consistent similarity graph learning that automatically integrates multiple view-specific graphs at the sample level, which can be formulated as follows:
\begin{equation}\label{3.2_7}
    \begin{aligned}
    \min_{\bm{\Omega}_2}&\sum_{v=1}^{V}\sum_{j=1}^{n}\!\|\bm{Z}_{. j}^{T}-\bm{q}_{\cdot j}^T \tilde{\bm{S}_j}\|_{2}^{2}\!+\alpha\operatorname{Tr}(\bm{F} \bm{L}_Z \bm{F}^T)+\eta \|\bm{Z}\|_{F}^{2}\\
    & +\beta\operatorname{Tr}({\bm{\Lambda}}^{(v)} {\bm{X}}^{(v)} \bm{L}_{S^{(v)}} ({\bm{\Lambda}}^{(v)} {\bm{X}}^{(v)})^T)+{\gamma}^{(v)}\|{\bm{S}}^{(v)}\|_{F}^{2}\\
    \text{s.t. }&\bm{Z}_{\cdot j} \!\!\geq\!\! 0, \bm{Z}_{\cdot j}^{T} \bm{1}_n\!\!=\!\!1,\bm{q}_{\cdot j}\!\!\geq\!\! 0, \bm{q}_{\cdot j}^T \bm{1}_V\!\!=\!\!1, \bm{S}_{\cdot j}^{(v)}\!\!\geq\!\! 0, {\bm{S}_{\cdot j}^{(v)}}^{T}\!\!\!\bm{1}_n\!\!=\!\!1,
    \end{aligned}
\end{equation}
where $\bm{\Omega}_2=\{\bm{Z},\bm{q}_{\cdot j},\bm{F},{\bm{\Lambda}}^{(v)},{{\bm{S}}^{(v)}}\}$. $\bm{Z} \in \mathbb{R}^{n \times n}$ is the consistent similarity graph, $\bm{S}^{(v)}$ denotes the similarity matrix for the $v$-th view, $\bm{L}_Z$ and $\bm{L}_{S^{(v)}}$ represent the Laplacian matrices of $\bm{Z}$ and $\bm{S}^{(v)}$, respectively, and $\alpha$, $\eta$, $\beta$ and ${\gamma}^{(v)}$ are regularization parameters. In Eq.~(\ref{3.2_7}), to obtain a consistent similarity graph with sample-level view weighting, we first stack the similarity matrices from all views into a third-order tensor $\tilde{\bm{S}} \in \mathbb{R}^{V \times n \times n}$, where the $j$-th slice $\tilde{\bm{S}_j} \in \mathbb{R}^{V \times n }$ represents the similarities between sample $j$ and all other samples across different views. Furthermore, we introduce a weight vector $\bm{q}_{\cdot j} \in \mathbb{R}^{V \times 1 }$ for each sample, where ${q}_{v j}$ denotes the weight of the $v$-th view for sample $j$. Then, the first term in Eq.~(\ref{3.2_7}) fuses the slices of $\tilde{\bm{S}}$  using automatically assigned sample-level weights to obtain the consistent similarity graph $\bm{Z}$, which better preserves the intrinsic local structure of the data compared to using the same fusion weights for all samples. In addition, the second term $\operatorname{Tr}(\bm{F} \bm{L}_Z \bm{F}^T)$ enforces that similar samples are assigned similar cluster labels, which helps to preserve the local structure in the label space. Furthermore, the fourth term performs adaptive similarity graph learning for each view within the selected feature space, ensuring that samples similar in the high-dimensional space remain similar in the low-dimensional feature space. This promotes the selection of discriminative features.

By integrating Eq.~(\ref{3.2_4}) and Eq.~(\ref{3.2_7}), we obtain the mathematical optimization model for KAFUSE as follows:
\begin{equation}\label{3.2_8}	
	\begin{aligned}
		\min_{\bm{\Omega}_{3}}&\!\!\sum_{v=1}^{V}\!\bm{\theta}_v^2\| {\bm{W}^{(v)}}^T\!\!\!\! \bm{\Lambda}^{(v)}\!\! \bm{X}^{(v)} \!\!\!+ \!\bm{b}^{(v)}\!\bm{1}^T_n\!\!\!- \!\!\bm{F} \|_F^2\!\!-\!\bm{\omega}_v^r\!\operatorname{Tr}(\!\bm{H}\!\bm{K}_c^{(\!v\!)} \!\!\bm{H} \!\bm{K}_{u}^{(\!v\!)}\!)\\
        &+\sum_{j=1}^{n}\|\bm{Z}_{. j}^{T}-\bm{q}_{\cdot j}^T \tilde{\bm{S}_j}\|_{2}^{2}+\alpha\operatorname{Tr}(\bm{F} \bm{L}_Z \bm{F}^T)+\eta \|\bm{Z}\|_{F}^{2}\\[-1pt]
		& +\beta\operatorname{Tr}({\bm{\Lambda}}^{(v)} {\bm{X}}^{(v)} \bm{L}_{S^{(v)}} ({\bm{\Lambda}}^{(v)} {\bm{X}}^{(v)})^T)+ {\gamma}^{(v)}\|{\bm{S}}^{(v)}\|_{F}^{2}\\
		\text{s.t. }&{\bm{W}^{(v)}}^T \!\bm{W}^{(v)} = \bm{I}_c,~\!\bm{F}\bm{F}^T =\bm{I}_c,~\!\bm{Z}_{\cdot j} \geq 0, \bm{Z}_{\cdot j}^{T} \bm{1}_n=1,\\
		&\bm{q}_{\cdot j}\!\geq\! 0, \bm{q}_{\cdot j}^T \bm{1}_V = 1, \bm{S}_{\cdot j}^{(v)} \!\geq \!0, {\bm{S}_{\cdot j}^{(v)}}^{T} \!\!\bm{1}_n\!=\!1,\bm{\lambda}^{(v)}\!\!\in\!\{0,\!1\},\\
		& \bm{\theta}_v \ge 0,~\bm{\omega}_v \ge 0,~\sum_{v=1}^{V}\bm{\theta}_v\!=\!1,~\sum_{v=1}^{V}\bm{\omega}_v\!=\!1,
	\end{aligned}
\end{equation}
where $\bm{\Omega}_{3}=\{\bm{W}^{(v)},\bm{\Lambda}^{(v)},\bm{b}^{(v)},\bm{F},\bm{Z},\bm{q}_{\cdot j},\bm{S}^{(v)},\bm{\theta}_v,\bm{\omega}_v\}$. 

Based on the Karush-Kuhn-Tucker (KKT) theorem~\cite{KuhnTucker1951}, the optimal value of $\bm{b}^{(v)}$ in Eq.~(\ref{3.2_8})  is attained at the point where the first-order derivative of its Lagrangian function equals zero. Accordingly, we define the Lagrangian function with respect to $\bm{b}^{(v)}$ as $\mathcal{L}(\bm{b}^{(v)}) \!\!= \!\!\|{\bm{W}^{(v)}}^T {\bm{\Lambda}}^{(v)} \bm{X}^{(v)} + \bm{b}^{(v)} \bm{1}_n^T - \bm{F}\|_F^2$. By differentiating $\mathcal{L}(\bm{b}^{(v)})$ w.r.t $\bm{b}^{(v)}$ and setting the result to zero, we obtain the optimal solution for $\bm{b}^{(v)}$ as follows:
\begin{equation}\label{4_1}
	\frac{\partial \mathcal{L}(\bm{b}^{(v)})}{\partial \bm{b}^{(v)}}=0 \Rightarrow \bm{b}^{(v)} = \frac{1}{n} ( \bm{F} \bm{1}_n - {\bm{W}^{(v)}}^T \bm{\Lambda}^{(v)} \bm{X}^{(v)} \bm{1}_n ).
\end{equation}

Substituting the optimal solution of $\bm{b}^{(v)}$ into Eq.~(\ref{3.2_8}), Eq.~(\ref{3.2_8}) can be rewritten into 
\begin{equation}\label{4_2}	
	\begin{aligned}
		\min_{\bm{\Omega}}&\!\sum_{v=1}^{V}\bm{\theta}_v^2\| {\bm{W}^{(v)}}^T \!\!\bm{\Lambda}^{(v)} \!\bm{X}^{(v)}\! \bm{H}\!\!-\!\!\bm{F} \bm{H} \|_F^2\!-\!\bm{\omega}_v^r\!\operatorname{Tr}(\!\bm{H}\!\bm{K}_c^{(v)} \!\!\bm{H} \!\bm{K}_{u}^{(v)}\!)\\
        &+\sum_{j=1}^{n}\|\bm{Z}_{. j}^{T}-\bm{q}_{\cdot j}^T \tilde{\bm{S}_j}\|_{2}^{2}+\eta \|\bm{Z}\|_{F}^{2}+\alpha\operatorname{Tr}(\bm{F} \bm{L}_Z \bm{F}^T)\\
		&+\beta\operatorname{Tr}({\bm{\Lambda}}^{(v)} {\bm{X}}^{(v)} \bm{L}_{S^{(v)}} ({\bm{\Lambda}}^{(v)} {\bm{X}}^{(v)})^T)+ {\gamma}^{(v)}\|{\bm{S}}^{(v)}\|_{F}^{2}\\
        &\text{s.t. the same constraints as in Eq.~(\ref{3.2_8}),}
	\end{aligned}
\end{equation}
where $\bm{\Omega}=\{\bm{W}^{(v)},\bm{\Lambda}^{(v)},\bm{F},\bm{Z},\bm{q}_{\cdot j},\bm{S}^{(v)},\bm{\theta}_v,\bm{\omega}_v\}$. The constraints for Eq.~(\ref{4_2}) are identical to those in Eq.~(\ref{3.2_8}), and are therefore omitted for brevity. There are five parameters in Eq.~(\ref{4_2}): $\alpha$, $\beta$, $\eta$, ${\gamma}^{(v)}$ and $r$. Among them, the regularization parameter $\eta$ and ${\gamma}^{(v)}$ are automatically determined during the optimization procedure, as described in subsection~\ref{3.D}, so only three parameters need to be manually tuned in the proposed KAFUSE method. There are two major advantages to the proposed KAFUSE method. On the one hand, it employs kernel alignment in conjunction with an orthogonal constraint to reduce feature redundancy in both linear and nonlinear manners, thereby facilitating the selection of features that are both discriminative and uncorrelated. On the other hand, unlike traditional methods that assign the same view weight to all samples, KAFUSE automatically adjusts the view weights for each individual sample. By assigning sample-level view weights during the consistent graph learning process, KAFUSE learns a more reliable similarity graph, which in turn better preserves the local geometric structure of the data.

\subsection{Optimization and Algorithm}\label{3.D}
Since the objective function in Eq.~(\ref{4_2}) is not jointly convex with respect to all variables, we propose an alternative iterative algorithm to solve the optimization problem. In each iteration, the algorithm updates one variable while keeping the others fixed.

\subsubsection{Update $\bm{W}^{(v)}$ by Fixing Other Variables}
When other variables are fixed, $\bm{W}^{(v)}$ can be updated by solving the following optimization problem:
\begin{equation}\label{4.1_1}
	\begin{aligned}
	&\min_{\bm{W}^{(v)}}\sum_{v=1}^{V}\bm{\theta}_v^2\| {\bm{W}^{(v)}}^T \bm{\Lambda}^{(v)} \bm{X}^{(v)} \bm{H} - \bm{F} \bm{H} \|_F^2\\
	&\text{ s.t. }~ {\bm{W}^{(v)}}^T \bm{W}^{(v)} = \bm{I}_c.
    \end{aligned}
\end{equation}

Since $\bm{H}\bm{H}\!=\!\bm{H}$ and $\bm{H}^T\!=\!\bm{H}$, Eq.~(\ref{4.1_1}) can be equivalently rewritten as follows:
\begin{equation}\label{4.1_2}
	\begin{aligned}
	&\min_{\bm{W}^{(v)}}\!\sum_{v=1}^{V}\bm{\theta}_v^2\operatorname{Tr}({\bm{W}^{(v)}}^T \bm{J}^{(v)} \bm{W}^{(v)}-2{\bm{W}^{(v)}}^T \bm{M}^{(v)})\\
	&\text{ s.t. }~ {\bm{W}^{(v)}}^T \bm{W}^{(v)} = \bm{I}_c,
	\end{aligned}
\end{equation}
where $\bm{J}^{(v)}\!\!=\!\!\bm{\Lambda}^{(v)}\!\bm{X}^{(v)}\!\bm{H} (\bm{\Lambda}^{(v)} \!\bm{X}^{(v)})^T$, $\bm{M}^{(v)}\!\!=\!\!\bm{\Lambda}^{(v)}\! \bm{X}^{(v)} \!\bm{H} \!\bm{F}^T$. Following~\cite{nie2017generalized}, Eq.~(\ref{4.1_2}) can be solved using the Generalized Power Iteration (GPI) method.

\subsubsection{Update $\bm{F}$ by Fixing Other Variables}
Fixing the other variables, we can update $\bm{F}$ by solving the following problem:
\begin{equation}\label{4.1_3}
	\begin{aligned}
	&\min_{\bm{F}}\!\sum_{v=1}^{V}{\!\bm\theta}_v^2\| {\bm{W}^{(v)}}^T \!\!\bm{\Lambda}^{(v)} \!\bm{X}^{(v)}\!\bm{H}\!\! - \!\!\bm{F}\! \bm{H} \|_F^2\!+\!\alpha\operatorname{Tr}(\bm{F} \!\bm{L}_Z \!\bm{F}^T)\\
	&\text{ s.t. }~\bm{F}\bm{F}^T\!= \bm{I}_c.
    \end{aligned}
\end{equation}

According to the properties of matrix norms, Eq.~(\ref{4.1_3}) can be equivalently rewritten as:
\begin{equation}\label{4.1_4}
	\begin{aligned}
	&\min_{\bm{F}}\operatorname{Tr}(\bm{F} \bm{G} \bm{F}^T -2 \bm{F} \bm{E}),\\
    &\text{ s.t. }~\bm{F}\bm{F}^T\!= \bm{I}_c.
    \end{aligned}
\end{equation}
where $\bm{G}=\sum_{v=1}^{V}(\alpha \bm{L}_Z - \bm{\theta}_v^2 \bm{H})$, and $\bm{E}=\sum_{v=1}^{V} \bm{\theta}_v^2 \bm{H}(\bm{\Lambda}^{(v)} \bm{X}^{(v)})^T \bm{W}^{(v)}$. Eq.~(\ref{4.1_4}) can also be solved using the Generalized Power Iteration (GPI) method, in the same way as for $\bm{W}^{(v)}$.

\subsubsection{Update $\bm{Z}$ by Fixing Other Variables}
With the other variables fixed, $\bm{Z}$ can be updated by solving the following problem:
\begin{equation}\label{4.1_5}
	\begin{aligned}
		\min_{\bm{Z}}&\alpha\operatorname{Tr}(\bm{F} \bm{L}_Z \bm{F}^T)+\sum_{j=1}^{n}\|\bm{Z}_{\cdot j}^{T}-\bm{q}_{\cdot j}^T \tilde{\bm{S}_j}\|_{2}^{2} + \eta \|\bm{Z}\|_{F}^{2}\\
		\text{ s.t. }&\bm{Z}_{\cdot j} \geq 0, \bm{Z}_{\cdot j}^{T} \bm{1}_n=1.
	\end{aligned}
\end{equation}
Since each column of $\bm{Z}$ in Eq.~(\ref{4.1_5}) is independent, $\bm{Z}$ can be optimized by solving for each column $\bm{Z}_{\cdot j}$ individually. The optimization problem for $\bm{Z}_{\cdot j}$ as follows:
\begin{equation}\label{4.1_6}
	\begin{aligned}
		\min_{\bm{Z}_{\cdot j}}& \alpha \sum_{j=1}^{n}\bm{Z}_{\cdot j}^{T}\bm{D}_{\cdot j}\!+\!\sum_{j=1}^{n}\|\bm{Z}_{\cdot j}^{T}\!-\!\bm{q}_{\cdot j}^T \tilde{\bm{S}_j}\|_{2}^{2} +\! \eta \!\sum_{j=1}^{n}\!\bm{Z}_{\cdot j}^{T}\bm{Z}_{\cdot j} \\
		\text{ s.t. }&\bm{Z}_{\cdot j} \geq 0, \bm{Z}_{\cdot j}^{T} \bm{1}_n=1.
		\end{aligned}
\end{equation}
where the $i$-th entry of the vector $\bm{D}_{\cdot j}$ is $\bm{D}_{i j}=\frac{1}{2}\|\bm{F}_{\cdot i}-\bm{F}_{\cdot j}\|_{2}^{2}$. Then, we construct the Lagrangian function with respect to $\bm{Z}_{\cdot j}$ as follows:
\begin{equation}\label{4.1_7}
	\mathcal{L}(\bm{Z}_{\cdot j})=\frac{1}{2}\|\bm{Z}_{\cdot j}+ \frac{\bm{A}_{\cdot j}}{2(1+\eta)}\|_{F}^{2}-\mu(\bm{Z}_{\cdot j}^{T} \bm{1}_n-1)-\bm{\rho}^T\bm{Z}_{\cdot j},
\end{equation}
where $\bm{A}_{. j}\!\!=\!\alpha\bm{D}_{\cdot j}-2(\tilde{\bm{S}_j}^T \bm{q}_{\cdot j})$, and $\mu$ and $\bm{\rho}$ are the Lagrangian multipliers. By taking the derivative of the Lagrangian  $\mathcal{L}(\bm{Z}_{\cdot j})$ with respect to $\bm{Z}_{\cdot j}$ and applying  the KKT condition $\rho_i \bm{Z}_{ij} = 0$, we have:
\begin{equation}\label{4.1_Z1}
	\bm{Z}_{\cdot j}=max(\mu \bm{1}_n - \frac{\bm{A}_{\cdot j}}{2(1+\eta)},0),
\end{equation}
where $max(x,0)$ returns the larger of $x$ and 0.

In practical applications, we usually focus on neighboring samples that are highly similar to a given data point. Specifically, $\bm{Z}_{\cdot j}$ is constrained to have only $k$ nonzero elements, where $k$ represents the number of neighboring data points. Additionally, the entries of $\bm{A}_{\cdot j}$ are sorted in ascending order. Accordingly, $\bm{Z}_{\cdot j}$ must satisfies $\bm{Z}_{k,j} > 0$ and $\bm{Z}_{k+1,j} = 0$, which yields:
\begin{equation}\label{4.1_Z1}
	\mu - \frac{\bm{A}_{k,j}}{2(1+\eta)}> 0,~\mu - \frac{\bm{A}_{k+1,j}}{2(1+\eta)} \leq 0.
\end{equation}
Based on the constraint $\bm{Z}_{\cdot j}^{T}\bm{1}_n = 1$, the Lagrange multiplier $\mu$ can be calculated as $\frac{1}{k}+\frac{1}{2k(1+\eta)}\sum_{i=1}^{k}\bm{A}_{ij}$. Furthermore, by combining Eq.~(\ref{4.1_Z1}), the parameter $\eta$ can be determined as $(k \bm{A}_{k+1,j}-\sum_{i=1}^{k}{\bm{A}}_{i j})/{2}-1$. The optimal solution of $\bm{Z}$ is then given by:
\begin{equation}\label{4.1_7}
\bm{Z}_{ij}=\left\{\begin{array}{cc}
\frac{{\bm{A}}_{k+1,j}-\bm{A}_{i j}}{k{\bm{A}}_{k+1,j}-\sum_{i=1}^{k}{\bm{A}}_{i j}}& i \le k; \\
0 & i > k.
\end{array}\right.
\end{equation}

\subsubsection{Update $\bm{S}^{(v)}$ by Fixing Other Variables}
While  the other variables are fixed, the objective function with respect to $\bm{S}^{(v)}$ is reduced to:
\begin{equation}\label{4.1_8}
	\begin{aligned}
		\min_{\bm{S}^{(v)}}&\sum_{j=1}^{n}\beta\operatorname{Tr}({\bm{\Lambda}}^{(v)} {\bm{X}}^{(v)} \bm{L}_{S^{(v)}} ({\bm{\Lambda}}^{(v)} {\bm{X}}^{(v)})^T)\\
		&+\|\bm{Z}_{. j}^{T}\!-\!\bm{q}_{\cdot j}^T \tilde{\bm{S}_j}\|_{F}^{2}+{\gamma}^{(v)}\|{\bm{S}}^{(v)}\|_{F}^{2}\\
		\text{ s.t. }&~\bm{S}_{\cdot j}^{(v)} \geq 0, {\bm{S}_{\cdot j}^{(v)}}^{T} \bm{1}_n=1.
	\end{aligned}
\end{equation}
Since the columns of $\bm{S}^{(v)}$ are independent,  the optimization can be performed separately for each column $\bm{S}_{\cdot j}^{(v)}$. Accordingly, following a similar approach as used for optimizing $\bm{Z}$,  the optimal solution for
$\bm{S}_{\cdot j}^{(v)}$ is given by:
\begin{equation}\label{4.1_10}
\bm{S}_{ij}^{(v)}=\left\{\begin{array}{cc}
\frac{\bm{N}^{(v)}_{k+1,j}-\bm{N}^{(v)}_{i j}}{k\bm{N}^{(v)}_{k+1,j}-\sum_{i=1}^{k}\bm{N}^{(v)}_{i j}}& i \le k; \\
0 & i > k.
\end{array}\right.
\end{equation}
where $\bm{N}^{(v)}_{\cdot j}=\frac{\beta}{2} \bm{O}^{(v)}_{\cdot j}\!-\! 2{q}_{v j}\bm{Z}_{\cdot j}$, and the $i$-th entry of the vector $\bm{O}^{(v)}_{\cdot j}$ is $\bm{O}^{(v)}_{ij}\!\!=\!\!\|(\bm{\Lambda}^{(v)}\bm{X}^{(v)})_{\cdot i}-(\bm{\Lambda}^{(v)}\bm{X}^{(v)})_{\cdot j}\|_{2}^{2}$. The entries of $\bm{N}^{(v)}_{\cdot j}$ are sorted in ascending order. To ensure that $\bm{S}_{\cdot j}^{(v)}$ contains exactly $k$ nonzero entries, the regularization parameter ${\gamma}^{(v)}$ is set to $(k {\bm{N}^{(v)}}_{k+1,j}-\sum_{i=1}^{k}\bm{N}^{(v)}_{i j})/{2}-{q}_{v j}^{2}$.

\subsubsection{Update $\bm{q}_{\cdot j}$ by Fixing Other Variables}
With the remaining variables fixed, the optimization problem with respect to $\bm{q}_{\cdot j}$ can be formulated as follows:
\begin{equation}\label{4.1_12}
	\begin{aligned}
		\min_{\bm{q}_{\cdot j}}&\sum_{j=1}^{n}\!\|\bm{Z}_{. j}^{T}\!-\!\bm{q}_{\cdot j}^T \tilde{\bm{S}_j}\|_{2}^{2}\\
		\text{ s.t. }~ &\bm{q}_{\cdot j}\geq 0, \bm{q}_{\cdot j}^T \bm{1}_V=1,
	\end{aligned}
\end{equation}
Then, the Lagrangian function for problem (\ref{4.1_12}) is given by:
\begin{equation}\label{4.1_13}
    \mathcal{L}(\bm{q}_{\cdot j})=\bm{q}_{\cdot j}^T \bm{B}_{j} \bm{B}_{ j}^T \bm{q}_{\cdot j}+\xi(1-\bm{q}_{\cdot j}^T \bm{1}_V)-\bm{\delta}^T \bm{q}_{\cdot j},
\end{equation}
where $\bm{B}_j=\bm{1}_V \bm{Z}_{. j}^T-\tilde{\bm{S}_j}$, and $\xi$ and $\bm{\delta}$ are the Lagrange multipliers. By taking the derivative of Eq.~(\ref{4.1_13}) with respect to $\bm{q}_{\cdot j}$ and applying the KKT conditions $\delta_v q_{v j}=0$, the update rule for $\bm{q}_{\cdot j}$ is derived as:
\begin{equation}\label{4.1_14}
        \bm{q}_{\cdot j}=\frac{(\bm{B}_{j} \bm{B}_{j}^T)^{-1} \bm{1}_V}{\bm{1}_V^T (\bm{B}_{j} \bm{B}_{j}^T)^{-1} \bm{1}_V}.
\end{equation}

\subsubsection{Update $\bm{\theta}_v$ by Fixing Other Variables}
While the other variables are fixed, we can update $\bm{\theta}_v$ by solving the following problem:
\begin{equation}\label{4.1_15}
    \begin{aligned}
        &\min_{\bm{\theta}_v}\sum_{v=1}^{V}~\bm{\theta}_v^{2} ~g^{(v)}\\
        &\text{s.t. }\bm{\theta}_v \ge 0, \sum_{v=1}^{V}\bm{\theta}_v=1,
    \end{aligned}
\end{equation}
where $g^{(v)}= \| {\bm{W}^{(v)}}^T \bm{\Lambda}^{(v)} \bm{X}^{(v)} \bm{H} - \bm{F} \bm{H} \|_F^2$. The Lagrangian function for problem (\ref{4.1_15}) can be formulated as follows:
\begin{equation}\label{4.1_16}
    \mathcal{L}(\bm{\theta}_v)=\sum_{v=1}^{V} \bm{\theta}_v^{2}g^{(v)}-\phi(\sum_{v=1}^{V}\bm{\theta}_v-1),
\end{equation}
where $\phi$ is the Lagrange multiplier. By setting the derivative of Eq.~(\ref{4.1_16}) with respect to $\bm{\theta}_v$ to zero and enforcing the constraint $\sum_{v=1}^{V}\bm{\theta}_v=1$, we obtain the optimal solution for $\bm{\theta}_v$ as follows:
\begin{equation}\label{4.1_17}
    \bm{\theta}_v = \frac{(g^{(v)})^{-1}}{\sum_{v=1}^{V}(g^{(v)})^{-1}}.
\end{equation}

\subsubsection{Update $\bm{\omega}_v$ by Fixing Other Variables}
With the other variables fixed, we can update $\bm{\omega}_v$ by solving the following problem:
\begin{equation}\label{4_18}
    \begin{aligned}
        &\min_{\bm{\omega}_v}\sum_{v=1}^{V}~{\bm{\omega}_v^r}~ h^{(v)}\\
        &\text{s.t. }\bm{\omega}_v \ge 0, \sum_{v=1}^{V}\bm{\omega}_v=1,
    \end{aligned}   
\end{equation}
where $h^{(v)}=\operatorname{Tr}(\bm{H} \bm{K}_c^{(v)} \bm{H} \bm{K}_{u}^{(v)})$. Analogous to the procedure for solving the $\bm{\theta}_v$ subproblem, the optimal solution for $\bm{\omega}_v$ can be obtained as:
\begin{equation}\label{4_19}
    \bm{\omega}_v = \frac{(h^{(v)})^{\frac{1}{1-r}}}{{\sum_{v=1}^{V}(h^{(v)})^{\frac{1}{1-r}}}}.
\end{equation}

\subsubsection{Update $\bm{\Lambda^{(v)}}$ by Fixing Other Variables}
While the other variables are fixed, the optimization problem for $\bm{\Lambda}^{(v)}$ is formulated as:
\begin{equation}\label{4.20}
	\begin{aligned}
	\min_{\bm{\Lambda}^{(v)}}&\!\sum_{v=1}^{V}\!\bm{\theta}_v^2\| {\bm{W}^{(v)}}^T \!\!\!\bm{\Lambda}^{(v)}\!\!\bm{X}^{(v)} \!\bm{H} \!\!- \!\!\bm{F}\!\bm{H} \|_F^2\!-\!\!
	\bm{\omega}_v^r \!\operatorname{Tr}(\!\bm{H} \!\bm{K}_c^{(v)}\! \bm{H} \!\bm{K}_{u}^{(v)}\!)\\
	&+\beta\operatorname{Tr}({\bm{\Lambda}}^{(v)} {\bm{X}}^{(v)} \bm{L}_{S^{(v)}} ({\bm{\Lambda}}^{(v)} {\bm{X}}^{(v)})^T)\\
	\text{s.t.}~&\bm{\lambda}^{(v)}\in \{0,1\}.
    \end{aligned}
\end{equation}
Following a commonly used relaxation strategy~\cite{gu2011towards,Theta2014robust}, the discrete variables can be relaxed to continuous ones, and the relaxed objective function is given as follows:
\begin{equation}\label{4.21}
	\begin{aligned}
	\min_{\bm{\Lambda}^{(v)}}&\!\sum_{v=1}^{V}\!\bm{\theta}_v^2 (\!{\bm{\lambda}^{(v)}}^T \!\bm{U}^{(v)} \!\bm{\lambda}^{(v)}\!\! -\!\! {{\bm{\lambda}}^{(v)}}^T \!\!\bm{e}^{(v)}\!)\!\!-\!\!\bm{\omega}_v^r \!\operatorname{Tr}(\!\bm{H} \!\bm{K}_c^{(v)}\!\bm{H} \!\bm{K}_{u}^{(v)}\!)\\
	&+\beta\operatorname{Tr}({\bm{\Lambda}}^{(v)} {\bm{X}}^{(v)} \bm{L}_{S^{(v)}} ({\bm{\Lambda}}^{(v)} {\bm{X}}^{(v)})^T)+\zeta\|\bm{\Lambda}^{(v)}\|_1\\
    \text{s.t.}~&\bm{\lambda}^{(v)}\in [0,1],
    \end{aligned}
\end{equation}
where $\bm{U}^{(v)}=(\bm{X}^{(v)} \bm{H}^T {\bm{X}^{(v)}}^T) \circ (\bm{W}^{(v)} {\bm{W}^{(v)}}^T)$, $\bm{e}^{(v)}=\operatorname{diag}(2 \bm{X}^{(v)} \bm{H} \bm{F}^T {\bm{W}^{(v)}}^T)$, and $\circ$ denotes the element-wise product. Eq.~(\ref{4.21}) can be solved by applying the proximal gradient descent method~\cite{sprechmann2015learning}:
\begin{equation}\label{4.22}
	{\bm{\Lambda}}^{(v)}=\bm{S}_{\Lambda t}(\bm{\Lambda}^{(v)} -t\nabla({\bm{\Lambda}}^{(v)})), \\
\end{equation}
where $t$ is the step size,
$\bm{S}_{\Lambda t}(\cdot)=\operatorname{sign}{(\cdot)}\max(|\cdot|-t,0)$ is the soft thresholding function, and $\nabla({\bm{\Lambda}}^{(v)})\!\!=\!\!\operatorname{diag}[\bm{\theta}_v^2((\bm{U}^{(v)}+{\bm{U}^{(v)}}^T)\bm{\lambda}^{(v)}-\bm{e}^{(v)})+\bm{\omega}_v^r\sum_{i,j=1}^{n}[((\bm{H} \bm{K}_{u}^{(v)} \bm{H}) \circ \bm{K}_{c}^{(v)})_{ij} (\bm{X}_{\cdot i}-\bm{X}_{\cdot j})^2]\frac{2 {\lambda}^{(v)}}{\sigma^2}-\bm{\omega}_v^r\sum_{i,j=1}^{n}[((\bm{H} \bm{K}_{c}^{(v)} \bm{H}) \circ \bm{K}_{u}^{(v)})_{ij} (\bm{X}_{\cdot i}-\bm{X}_{\cdot j})^2]\frac{2 (1-{\lambda}^{(v)})}{\sigma^2}+2 \beta {\bm{X}}^{(v)} {\bm{L}_{S^{(v)}}} ({\bm{X}}^{(v)})^T {\bm{\lambda}}^{(v)} ]$. 

The overall optimization procedure for KAFUSE is summarized in Algorithm 1. In the algorithm, $\bm{Z}$ is initialized as a matrix with all entries equal to $1/n$. The vectors $\bm{\theta}_v$, $\bm{\omega}_v$, and $\bm{q}_{\cdot j}$ are initialized so that each entry equals $1/V$, while the diagonal elements of $\bm{\Lambda}^{(v)}$ are set to $1/d_v$. Both $\bm{F}$ and $\bm{W}^{(v)}$ are randomly initialized as orthogonal matrices. In addition, $\bm{S}^{(v)}$  is initialized by constructing a $k$-nearest neighbor similarity matrix following \cite{ZS2019gmc}.

\subsection{Convergence and Complexity Analysis}
\subsubsection{Convergence Analysis} \label{prove-converge}
The objective function consists of multiple terms involving variables 
$\bm{W}^{(v)}$, $\bm{F}$, $\bm{Z}$, $\bm{S}^{(v)}$, $\bm{q}_{\cdot j}$, $\bm{\theta}_v$, $\bm{\omega}_v$ and ${\bm{\Lambda}}^{(v)}$. As the overall optimization problem is non-convex, we employ an alternating optimization strategy that decomposes the objective into eight subproblems, as outlined in Algorithm 1. The convergence of Algorithm 1 can be established by demonstrating that each subproblem converges individually. Specifically, the subproblems for  $\bm{q}_{\cdot j}$, $\bm{\theta}_v$, $\bm{\omega}_v$, $\bm{S}^{(v)}$ and $\bm{Z}$ admit closed-form solutions, guaranteeing the convergence of their respective updates. Additionally, the subproblems involving $\bm{W}^{(v)}$ and $\bm{F}$ are solved using the GPI algorithm, whose convergence has been theoretically proven in~\cite{nie2017generalized}. The convergence of the ${\bm{\Lambda}}^{(v)}$ subproblem, solved via the proximal gradient descent algorithm, is ensured by the results in~\cite{ji2009accelerated}. Therefore, the proposed optimization algorithm is guaranteed to converge. Furthermore, empirical validation of the convergence of Algorithm 1 is provided in the experimental section.

\subsubsection{Complexity Analysis}
As shown in Algorithm 1, the variables are updated in an alternating manner. In each iteration, updating $\bm{W}^{(v)}$ and $\bm{F}$ using the GPI algorithm has computational complexities of $\mathcal{O}({d_v^2 n})$ and $\mathcal{O}({n^2 c})$, respectively. The updates for $\bm{Z}$ and $\bm{S}^{(v)}$ require $\mathcal{O}(n^2logn)$, while updating  ${\bm{\Lambda}}^{(v)}$ involves kernel computations with a time complexity of $\mathcal{O}(n^2 d_v)$. The updates for $\bm{q}_{\cdot j}$, $\bm{\theta}_v$ and $\bm{\omega}_v$ consist only of element-wise division operations, making their computational cost negligible. Therefore, the overall time complexity of Algorithm 1 per iteration is $\mathcal{O}(d^2n + n^2d)$, where $d=\sum_{v=1}^{V}d_v$.

\begin{algorithm}[t]
  \caption{Iterative algorithm of KAFUSE}
  \label{alg:algorithm}
  \textbf{Input}: The multi-view data $\{\bm{X}^{(v)}\}_{v=1}^V$, parameters $\alpha$, $\beta$, and $r$.
  \begin{algorithmic}[1]
\State \textbf{Initialize} $\{\bm{W}^{(v)}, \bm{\Lambda}^{(v)}, \bm{S}^{(v)}, \bm{\theta}_v, \bm{\omega}_v\}^V_{v=1}$, $\bm{F}$, $\bm{Z}$, and $\bm{q}_{\cdot j}$.
    \While{\textit{not convergent}}
        \State Update $\bm{W}^{(v)}$ according to Eq.~(\ref{4.1_2}).
        \State Update $\bm{F}$ according to Eq.~(\ref{4.1_4}).
        \State Update $\bm{Z}$ according to Eq.~(\ref{4.1_7}).
        \State Update $\bm{S}^{(v)}$ according to Eq.~(\ref{4.1_10}).
        \State Update $\bm{q}_{\cdot j}$ according to Eq.~(\ref{4.1_14}).
		\State Update $\bm{\theta}_v$ according to Eq.~(\ref{4.1_17}).
        \State Update $\bm{\omega}_v$ according to Eq.~(\ref{4_19}).
        \State Update ${\bm{\Lambda}}^{(v)}$ according to Eq.~(\ref{4.22}).
    \EndWhile
  	\end{algorithmic}
	\textbf{Output}: Sort the diagonal entries of $\{\bm{\Lambda}^{(v)}\}_{v=1}^V$ in descending order and select the top $l$ features.
\end{algorithm}

\section{Experiments}
In this section, we evaluate the performance of KAFUSE by conducting extensive experiments on several real-world datasets and comparing it with a range of state-of-the-art methods.

\subsection{Experimental Schemes}

1) Datasets: We use eight real-world multi-view datasets to assess the effectiveness of our proposed method. The details of these datasets are provided below.

\textit{NGs}\footnote{https://lig-membres.imag.fr/grimal/data.html\label{NGs}} is comprised of 500 news article samples and provides three distinct views, each obtained using a different acquisition method.

\textit{Prokaryotic}\footnote{https://gitee.com/zhangfk/multi-view-dataset} is a dataset of prokaryotic phyla containing 551 species, each described by three heterogeneous views, including textual data and two distinct genomic representations.

\textit{COIL20}\footnote{https://www.cs.columbia.edu/CAVE/software/softlib/coil-20.php} consists of 1,440 grayscale images of 20 objects, with each object represented by 72 images. The dataset offers three distinct views based on LBP, HOG, and GIST features.

\textit{Cora}\footnotemark[\getrefnumber{NGs}] contains 2,708 documents across 7 classes, represented by 1,433 words in the content view and 2,708 citation links in the citation view.

\textit{AD}\footnote{https://archive.ics.uci.edu/ml/datasets/advertisement} consists of 3,279 examples representing advertisements on web pages. The features are derived from five views: phrases in URLs, image URLs, alt texts, anchor texts, and words occurring near the anchor text.

\textit{CiteSeer}\footnotemark[\getrefnumber{NGs}] comprises 3,312 text samples divided into six classes, and offers two views: a content view with 3,703 dimensions and a sample-link view with 3,312 dimensions.

\textit{ALOI}\footnote{http://elki.dbs.ifi.lmu.de/wiki/DataSets/MultiView \label{ALOI}} consists of 7,104 images from 64 classes and provides three distinct views, namely HSV color histograms, color similarity, and RGB color histograms.

\textit{Caltech}\footnote{\label{Caltech}\url{https://www.vision.caltech.edu/datasets}} contains 15,039 images spanning 90 object categories. Each image was collected from Google Images and is described by three feature views: LBP, GIST, and HOG.

The detailed statistics for these multi-view datasets are presented in Table~\ref{Table1}.

\begin{table}[!t]
\centering
\setlength{\tabcolsep}{1.05mm}
\caption{A detail description of datasets}\label{Table1}
\small
	\begin{tabular}{@{\extracolsep{\fill}}lcccc}
    	\toprule
		Datasets &\ Views & \ Samples & \ Features & \ Classes\\
		\midrule
		NGs&3&500&2000/2000/2000&5 \\
		Prokaryotic&3&551&393/3/438&4\\
		COIL20&3&1440&30/19/30&20\\
		Cora&2&2708&2708/1433&7\\
		AD&5&3279&457/495/472/19/111&2\\
		CiteSeer&2&3312&3312/3703&6\\
		ALOI&3&7104&77/64/64&64\\
		Caltech&3&15039&1024/512/2048&90\\
		\toprule
    \end{tabular}
\end{table}

2) Evaluation Metrics: Following previous studies~\cite{MFSGL2021autoweighted,MAMFS2021multilevel}, we apply the k-means clustering algorithm to the selected features to evaluate the performance of the proposed KAFUSE. To assess the quality of the features selected by different methods, we employ two widely used clustering metrics: Clustering Accuracy (ACC) and Normalized Mutual Information (NMI), which are defined  below.

\begin{equation}\label{ACC}
\text{ACC}=\frac{\sum_{i=1}^{n}\delta(y_i,map(\hat{y}_i))}{n},
\end{equation}
where $n$ denotes the number of samples, $y_i$ is the true label of the $i$-th sample, $\hat{y}_i$ is its assigned cluster label, $map(\hat{y}_i)$ is a function that maps the cluster label to the corresponding true label,  and $\delta$ is the Kronecker delta function, which returns 1 if its two arguments are equal and 0 otherwise.

\begin{equation}\label{NMI}
\text{NMI}=\frac{\sum_{i=1}^{p}\sum_{j=1}^{m}n_{ij}log(\frac{nn_{ij}}{n_{i}n_{j}})}{\sqrt{(\sum_{i=1}^{p}n_{i}log(\frac{n_{i}}{n}))(\sum_{j=1}^{m}n_{j}log(\frac{n_{j}}{n}))}},
	 \end{equation}
where $n$ is the total number of samples, $p$  denotes the number of ground truth classes, $m$ is the number of clusters, $n_{ij}$ represents the number of samples that belong to both the $i$-th ground-truth class and the $j$-th cluster, $n_{i}$ is the number of samples in the $i$-th class, and $n_{j}$ represents the number of samples in the $j$-th cluster.

3) Comparison Methods: To evaluate the effectiveness of the proposed approach, we compare KAFUSE with several state-of-the-art unsupervised feature selection methods. A brief overview of these methods is presented below:
\begin{itemize}
	\item \textbf{AllFeatures}  uses all the original features for comparison.
	\item \textbf{GAWFS}~\cite{2025GAWFS} integrates nonnegative matrix factorization with adaptive graph learning and employs a sparse feature weight matrix to automatically identify the most discriminative features.
	\item \textbf{RNE}~\cite{liu2020RNE} utilizes the locally linear embedding algorithm to construct the feature weight matrix and applies $\ell_{1}$-norm regularization to the loss function to select relevant features.
	\item \textbf{CFSMO}~\cite{CFSMO2024multi} constructs graphs that incorporate multi-order neighbor information and projects the data onto a shared latent representation to improve the selection of relevant features.
	\item \textbf{CDMvFS}~\cite{CDMvFS2024multi} generates mutually exclusive graphs to improve complementarity between views and integrating graph learning with consensus clustering to enforce consistency.
	\item \textbf{MAMFS}~\cite{MAMFS2021multilevel} utilizes adaptive $\mathit{k}\text{-nearest}$ neighbors to learn a collaborative similarity graph for feature selection.
	\item \textbf{CE-UMFS}~\cite{CE2024consistency} employs the Hilbert-Schmidt Independence Criterion together with a nuclear norm constraint to facilitate the selection of important features.
	\item \textbf{MFSGL}~\cite{MFSGL2021autoweighted} learns a consensus similarity graph across multiple views and incorporates a rank constraint to jointly optimize the similarity matrix and select features.
	\item \textbf{UKMFS}~\cite{2025UKMFS} uses binary hashing to generate weakly-supervised labels that guide feature selection.
	\item \textbf{WLTL}~\cite{2025WLTL} integrates multi-view spectral clustering with a weighted low-rank tensor framework to yield  pseudo-labels for feature selection.
\end{itemize}

To ensure a fair comparison, we set the parameter ranges for all baseline methods according to their original papers to achieve optimal performance. For our method, we employ  a grid search strategy to tune the parameters  $\alpha$ and $\beta$ within the range $\{10^{-3}, 10^{-2}, \ldots , 10^{3}\}$, and $r$  over $\{2, 3, \ldots , 9\}$. Since the optimal number of selected features for each dataset is difficult to determine~\cite{zhang2019feature, 2017challenge}, the feature selection ratio is varied from 10\% to 50\% with a step size of 10\%. Then, k-means clustering is performed 50 times on the selected features, and the mean and standard deviation of the results are reported.

\begin{table*}[!htbp]
	\centering
	\caption{Means and standard deviations (\%) of ACC for various methods across different datasets.}\label{ACC0.3}
	\resizebox{\textwidth}{!}{
		\renewcommand\tabcolsep{5pt}
		\begin{tabular}{lcccccccc}
			\toprule[1pt]
			\renewcommand{\arraystretch}{0.8}
			\diagbox[width=7em,height=2em]{Methods}{Datasets} &  NGs & Prokaryotic & COIL20 & Cora & AD & CiteSeer & ALOI & Caltech \\
			\midrule[1pt]
			KAFUSE & $\mathbf{23.86\pm1.08}$ & $\mathbf{68.97\pm2.10}$ & $\mathbf{99.04\pm3.08}$ & $\mathbf{37.77\pm1.09}$ & $\mathbf{87.06\pm1.83}$ & $\mathbf{40.48\pm4.88}$ & $\mathbf{51.98\pm2.38}$ & $\mathbf{75.74\pm2.24}$\\
			AllFeatures & $21.82$$\pm1.69 \bullet$ & $56.24$$\pm3.77 \bullet$ & $81.40$$\pm4.11 \bullet$ & $33.55$$\pm4.00 \bullet$ & $81.41$$\pm7.10 \bullet$ & $37.46$$\pm8.46 \bullet$ & $50.93$$\pm2.58 \circ$ & $43.35$$\pm3.66 \bullet$\\
            GAWFS & $21.05$$\pm0.78 \bullet$ & $56.53$$\pm4.71 \bullet$ & $75.27$$\pm4.01 \bullet$ & $31.99$$\pm2.00 \bullet$ & $82.02$$\pm1.63 \bullet$ & $21.25$$\pm1.64 \bullet$ & $42.86$$\pm1.85 \bullet$ & $68.03$$\pm1.04 \bullet$ \\
			RNE & $21.29$$\pm0.86 \bullet$ & $59.50$$\pm4.38 \bullet$ & $84.28$$\pm6.01 \bullet$ & $32.18$$\pm2.69 \bullet$ & $82.13$$\pm6.09 \bullet$ & $21.87$$\pm6.03 \bullet$ & $46.61$$\pm1.63 \bullet$ & $57.88$$\pm2.26 \bullet$\\
			CFSMO & $22.76$$\pm2.72 \circ$ & $58.24$$\pm2.83 \bullet$ & $80.02$$\pm4.72 \bullet$ & $34.55$$\pm0.21 \bullet$ & $80.08$$\pm7.77 \bullet$ & $21.38$$\pm8.66 \bullet$ & $48.83$$\pm2.43 \bullet$ & $70.46$$\pm2.78 \bullet$ \\
			CDMvFS & $21.88$$\pm2.06 \bullet$ & $56.61$$\pm2.79 \bullet$ & $93.13$$\pm4.71 \bullet$ & $31.45$$\pm0.52 \bullet$ & $83.50$$\pm5.99 \bullet$ & $21.63$$\pm7.26 \bullet$ & $42.73$$\pm2.68 \bullet$ & $72.98$$\pm1.88 \bullet$\\
			MAMFS & $21.55$$\pm1.55 \bullet$ & $48.66$$\pm2.40 \bullet$ & $75.11$$\pm1.48 \bullet$ & $32.22$$\pm3.62 \bullet$ & $82.97$$\pm6.60 \bullet$ & $24.11$$\pm2.42 \bullet$ & $45.97$$\pm2.11 \bullet$ & $59.37$$\pm2.54 \bullet$\\
			CE-UMFS & $21.26$$\pm0.92 \bullet$ & $56.48$$\pm4.01 \bullet$ & $76.54$$\pm2.25 \bullet$ & $31.52$$\pm3.55 \bullet$ & $86.01$$\pm3.28 \circ$ & $23.36$$\pm3.84 \bullet$ & $47.72$$\pm2.26 \bullet$ & $63.58$$\pm3.74 \bullet$ \\
			MFSGL & $21.61$$\pm0.61 \bullet$ & $53.76$$\pm5.91 \bullet$ & $84.12$$\pm4.71 \bullet$ & $30.44$$\pm1.97 \bullet$ & $85.99$$\pm1.63 \circ$ & $21.45$$\pm2.25 \bullet$ & $48.25$$\pm1.80 \bullet$ & $73.65$$\pm2.51 \bullet$ \\
			UKMFS & $21.40$$\pm1.13 \bullet$ & $59.39$$\pm1.61 \bullet$ & $93.60$$\pm5.61 \bullet$ & $33.95$$\pm3.55 \bullet$ & $82.02$$\pm1.32 \bullet$ & $36.38$$\pm3.16 \bullet$ & $50.09$$\pm2.54 \bullet$ & $50.38$$\pm2.11 \bullet$ \\
			WLTL & $21.44$$\pm0.64 \bullet$ & $59.76$$\pm6.21 \bullet$ & $97.20$$\pm5.79 \bullet$ & $35.14$$\pm4.64 \bullet$ & $84.77$$\pm7.67 \bullet$ & $38.69$$\pm8.62 \circ$ & $50.51$$\pm1.37 \circ$ & $45.48$$\pm2.52 \bullet$ \\
			\bottomrule[1pt]
		\end{tabular}
	}
\end{table*}

\begin{table*}[!htbp]
	\centering
	\caption{Means and standard deviations (\%) of NMI for various methods across different datasets.}\label{NMI0.3}
	\resizebox{\textwidth}{!}{
		\renewcommand\tabcolsep{5pt}
		\begin{tabular}{lcccccccc}
			\toprule[1pt]
			
			\renewcommand{\arraystretch}{0.8}
			\diagbox[width=7em,height=2em]{Methods}{Datasets} & NGs & Prokaryotic & COIL20 & Cora & AD & CiteSeer & ALOI & Caltech\\
			\midrule[1pt]
			KAFUSE & $\mathbf{4.32\pm1.03}$ & $\mathbf{39.76\pm1.19}$ & $\mathbf{99.68\pm1.17}$ & $\mathbf{16.11\pm1.16}$ & $\mathbf{7.94\pm0.74}$ & $\mathbf{16.72\pm4.91}$ & $\mathbf{71.83\pm1.20}$ & $\mathbf{85.89\pm2.27}$\\
			AllFeatures & $2.01\pm1.37 \bullet$ & $8.30\pm5.83 \bullet$ & $93.02\pm2.43 \bullet$ & $9.09\pm4.18 \bullet$ & $2.13\pm3.22 \bullet$ & $14.49\pm7.32 \bullet$ & $70.07\pm1.23 \bullet$ & $54.06\pm5.19 \bullet$ \\
            GAWFS & $1.33\pm0.82 \bullet$ & $0.93\pm3.58 \bullet$ & $86.54\pm2.25 \bullet$ & $3.39\pm2.36 \bullet$ & $0.27\pm0.60 \bullet$ & $0.22\pm1.10 \bullet$ & $58.84\pm0.82 \bullet$ & $81.72\pm1.04 \bullet$ \\
			RNE & $1.58\pm0.81 \bullet$ & $31.90\pm5.64 \bullet$ & $94.16\pm2.34 \bullet$ & $3.56\pm2.80 \bullet$ & $2.51\pm5.19 \bullet$ & $0.92\pm5.43 \bullet$ & $64.51\pm0.82 \bullet$ & $63.22\pm2.40 \bullet$ \\
			CFSMO & $2.80\pm2.52 \bullet$ & $29.53\pm4.79 \bullet$ & $89.93\pm2.02 \bullet$ & $12.50\pm0.21 \bullet$ & $4.09\pm3.81 \bullet$ & $0.58\pm7.07 \bullet$ & $66.12\pm1.25 \bullet$ & $84.81\pm0.80 \circ$ \\
			CDMvFS & $2.20\pm1.47 \bullet$ & $15.57\pm3.84 \bullet$ & $97.55\pm3.12 \bullet$ & $1.86\pm0.31 \bullet$ & $2.07\pm1.70 \bullet$ & $0.71\pm6.42 \bullet$ & $62.40\pm1.28 \bullet$ & $84.73\pm1.79 \circ$ \\
			MAMFS & $1.92\pm1.55 \bullet$ & $2.23\pm2.11 \bullet$ & $89.82\pm1.83 \bullet$ & $7.69\pm4.53 \bullet$ & $3.99\pm4.24 \bullet$ & $2.26\pm1.51 \bullet$ & $64.50\pm1.18 \bullet$ & $71.10\pm2.51 \bullet$\\
			CE-UMFS & $1.51\pm0.80 \bullet$ & $3.47\pm1.01 \bullet$ & $90.93\pm1.84 \bullet$ & $2.41\pm4.41 \bullet$ & $1.98\pm3.07 \bullet$ & $2.39\pm3.53 \bullet$ & $64.86\pm1.70 \bullet$ & $63.94\pm3.04 \bullet$ \\
			MFSGL & $2.00\pm0.60 \bullet$ & $9.13\pm2.07 \bullet$ & $94.06\pm3.12 \bullet$ & $6.08\pm2.13 \bullet$ & $3.53\pm4.37 \bullet$ & $0.57\pm1.74 \bullet$ & $67.29\pm1.13 \bullet$& $84.74\pm2.51 \circ$ \\
			UKMFS & $1.70\pm1.30 \bullet$ & $28.84\pm0.73 \bullet$ & $97.56\pm2.19 \bullet$ & $14.10\pm4.41 \bullet$ & $6.07\pm3.60 \bullet$ & $13.39\pm2.26 \bullet$ & $70.60\pm1.12 \circ$ & $58.79\pm2.02 \bullet$ \\
			WLTL & $1.72\pm0.62 \bullet$ & $32.90\pm5.72 \bullet$ & $99.07\pm2.40 \circ$ & $12.17\pm3.72 \bullet$ & $1.94\pm3.42 \bullet$ & $15.41\pm7.15 \circ$ & $70.35\pm0.72 \circ$ & $55.26\pm0.72 \bullet$ \\
			\bottomrule[1pt]
		\end{tabular}
	}
\end{table*}

\begin{figure*}[t]
  \centering
  \includegraphics[width=\textwidth]{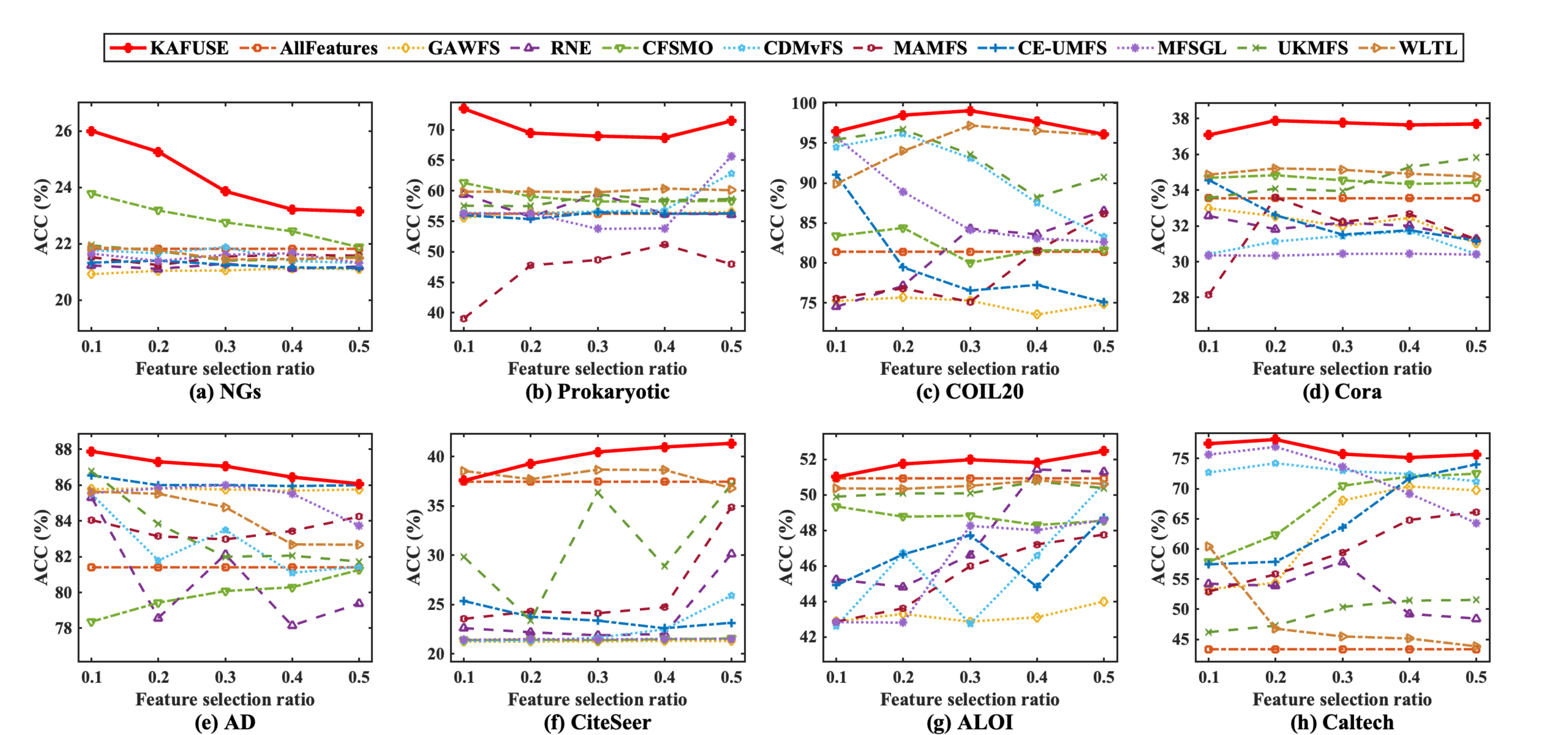}
  \caption{ACC of different methods under varying feature selection ratios.}
  \label{ACC8}
\end{figure*}

\begin{figure*}[t]
    \centering
    \includegraphics[width=\textwidth]{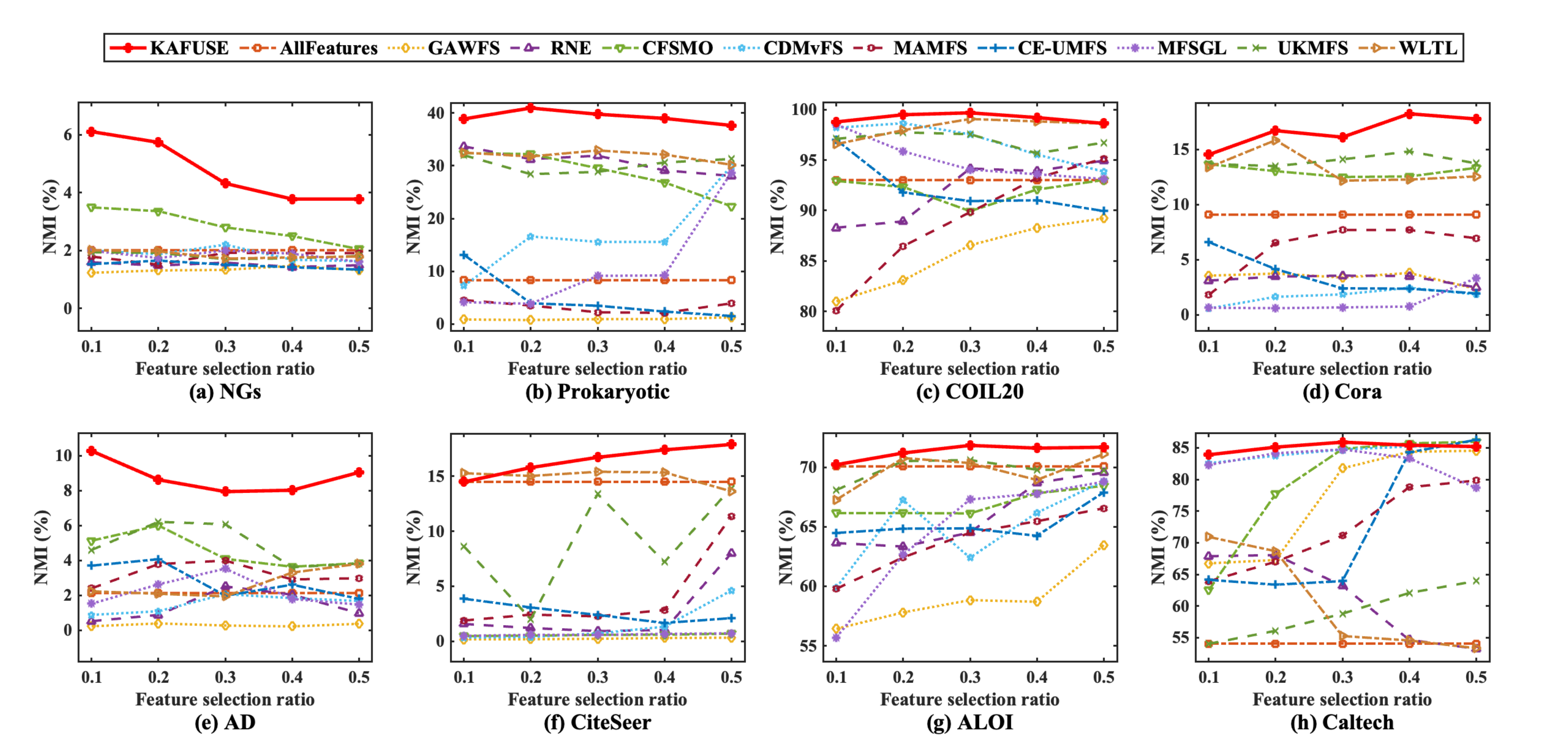}
    \caption{NMI of different methods under varying feature selection ratios.}
    \label{NMI8}
\end{figure*}

\subsection{Experimental Results}
In this subsection, we compare the clustering performance of the proposed KAFUSE with that of other competing methods in terms of ACC and NMI. Tables~\ref{ACC0.3} and~\ref{NMI0.3} respectively show the experimental results of ACC and NMI for different methods, with the feature selection ratio fixed at 30\%. The best results are shown in bold. Furthermore, we employ the Wilcoxon rank-sum test~\cite{2007wilcoxon} to assess whether the proposed KAFUSE demonstrates a statistically significant improvement over other comparative methods. At a significance level of 0.05, values marked with $\bullet$ in Tables~\ref{ACC0.3} and \ref{NMI0.3} indicate that KAFUSE is statistically superior to the other methods, while the symbol $\circ$ denotes that there is no statistically significant difference between KAFUSE and the comparative methods.

As shown in Tables~\ref{ACC0.3} and~\ref{NMI0.3}, our method KAFUSE consistently achieves better results than the other comparison methods in most cases. For the Prokaryotic, CiteSeer, and Caltech datasets, KAFUSE demonstrates average improvements of over 11\% in both ACC and NMI compared to other competing methods. On the COIL20, Cora, and ALOI datasets, KAFUSE gains average improvements of more than 4\% in ACC and 5\% in NMI compared to other methods. On the NGs and AD datasets, the proposed method maintains its performance advantage achieving average improvements approaching 2\% in both ACC and NMI. Furthermore, KAFUSE outperforms AllFeatures across all datasets in both ACC and NMI, demonstrating the effectiveness of the proposed method. We can also see that KAFUSE consistently outperforms all single-view-based methods, demonstrating the effectiveness of our cross-view consistent graph learning over the simple stacking of multi-view data into a single-view format used by single-view-based methods.

Since determining the optimal number of selected features for each dataset is challenging, we compared the performance of the different methods across a range of feature selection ratios. Figs.~\ref{ACC8} and~\ref{NMI8} illustrate the performance of various methods in terms of ACC and NMI, respectively, across feature selection ratios ranging from 10\% to 50\%. As shown in the two figures, KAFUSE consistently performs better than the other baseline methods in most cases. The superior performance of the proposed method can be attributed to its ability to reduce feature redundancy in both linear and nonlinear ways, as well as to adaptively learn a cross-view consistent similarity graph at the sample level, thereby enabling the selection of representative features.

\begin{figure*}[t]
	\centering
	\includegraphics[width=\textwidth, keepaspectratio]{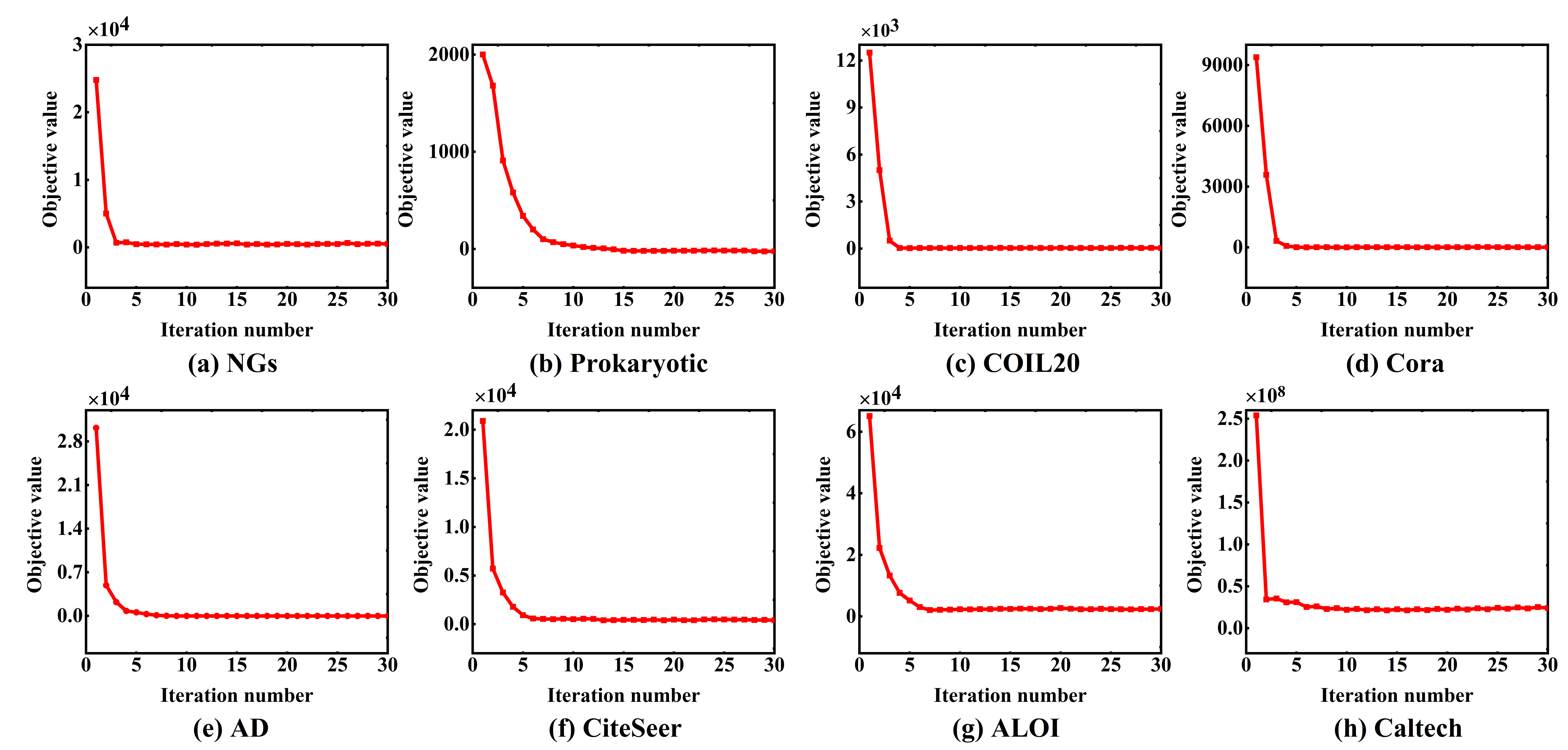}\caption{Convergence curves of KAFUSE on eight datasets.}
	\label{Convergence2}
\end{figure*}

\begin{figure*}[t]
	\centering
	\includegraphics[width=0.95\textwidth, keepaspectratio]{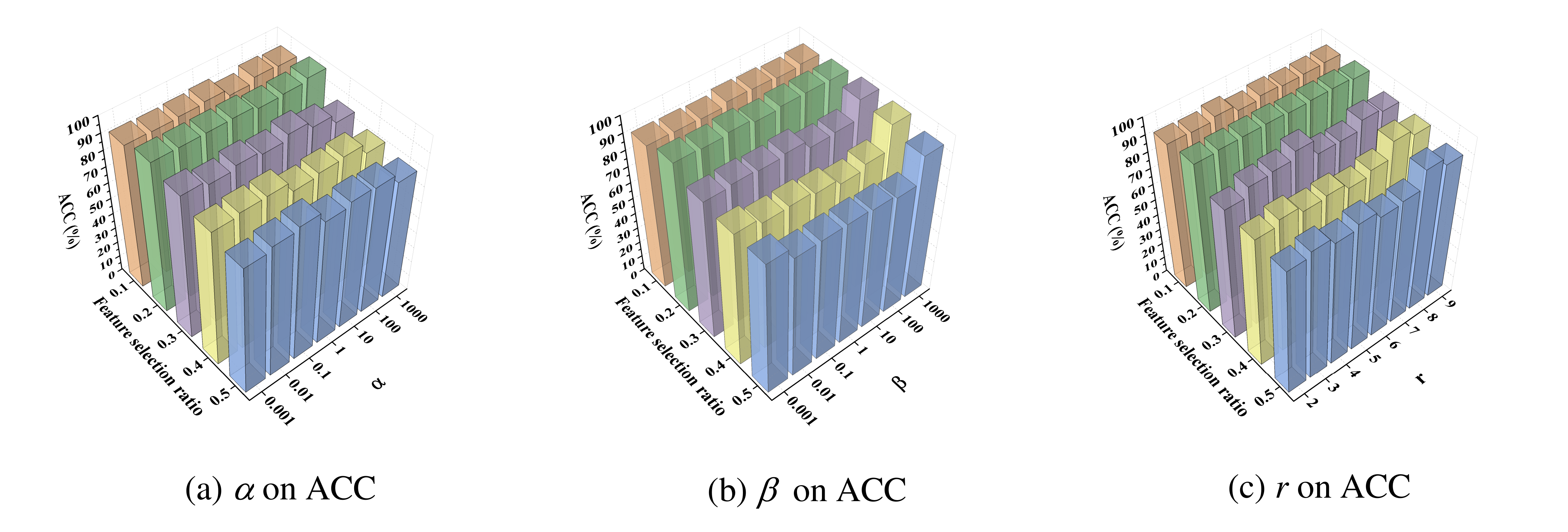}
	\caption{ACC of KAFUSE with varying parameters $\alpha$, $\beta$ and $r$ and feature selection ratios on COIL20 dataset.}
	\label{ACCSensitivity}
\end{figure*}

\begin{figure*}[!t]
	\centering
	\includegraphics[width=0.95\textwidth, keepaspectratio]{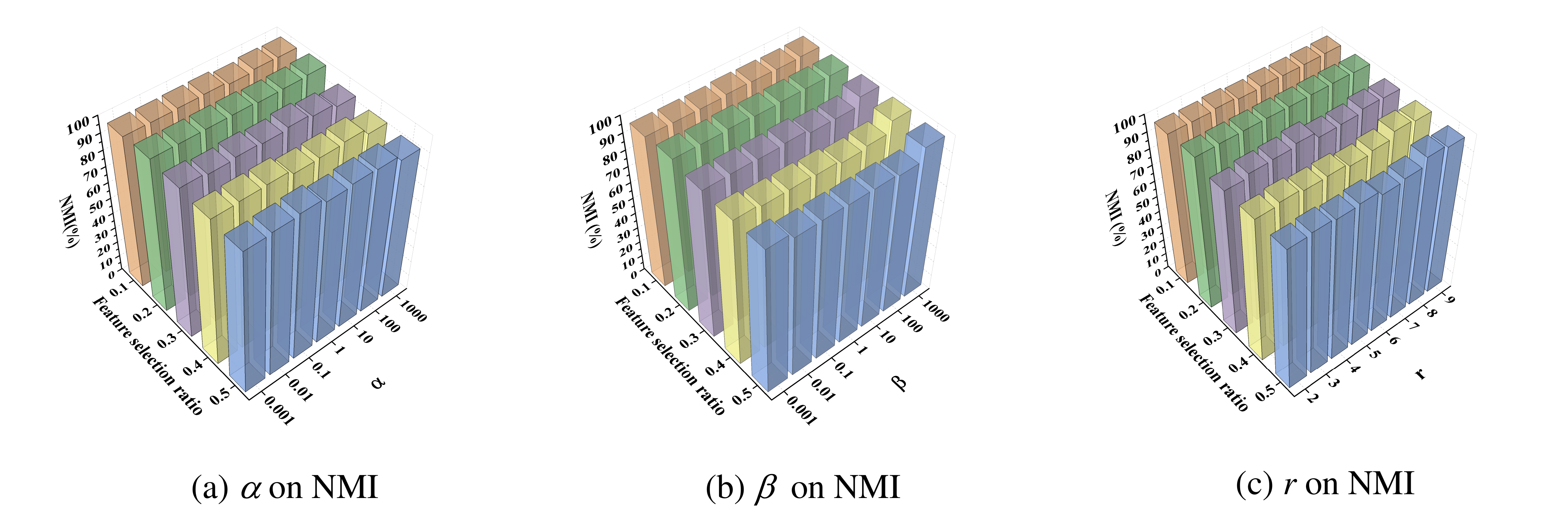}
	\caption{NMI of KAFUSE with varying parameters $\alpha$, $\beta$ and $r$ and feature selection ratios on COIL20 dataset.}
	\label{NMISensitivity}
\end{figure*}

\begin{figure}[h]
    \captionsetup{justification=raggedright, singlelinecheck=false}
    \centering
    \includegraphics[width=\linewidth]{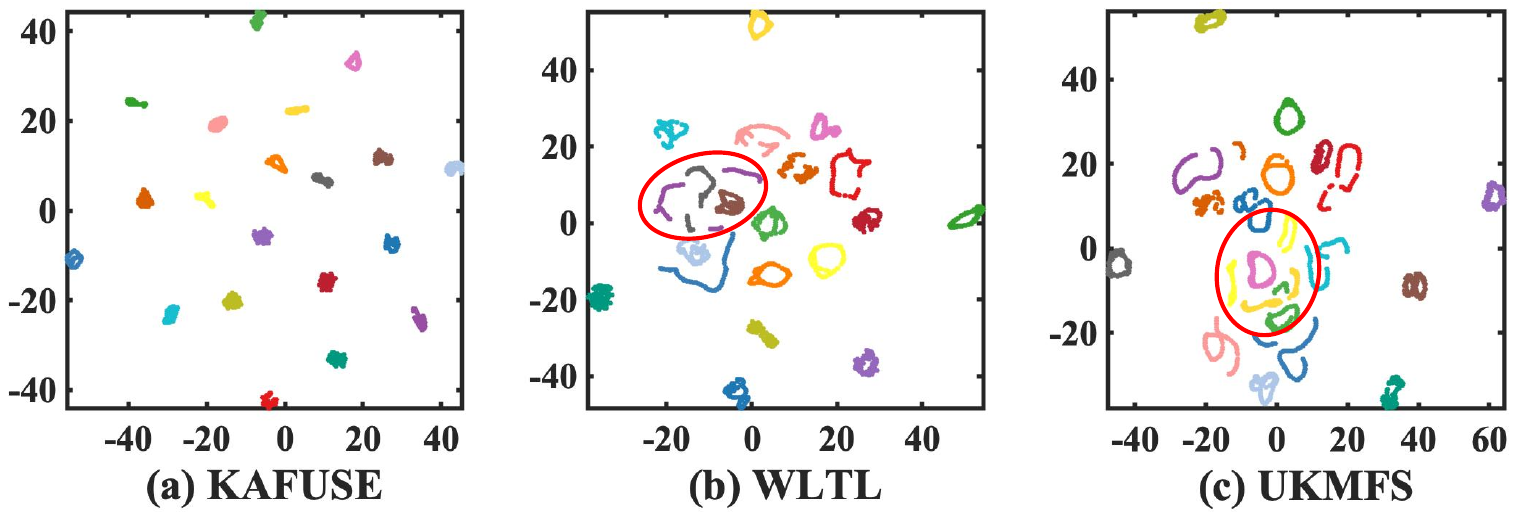}
    \caption{t-SNE visualizations of the top three performing methods on COIL20 dataset.}
    \label{tsne}
\end{figure}

\subsection{Convergence Behavior and Parameter Sensitivity}
In this subsection, we analyze the convergence and parameter sensitivity of the KAFUSE method. Fig.~\ref{Convergence2} shows how the objective value varies with the number of iterations on eight datasets. As illustrated in this figure, the convergence curve of KAFUSE decreases sharply during the first 5 iterations and becomes stable after about 20 iterations.

As shown in Eq.~(\ref{4_2}), our method involves three parameters: $\alpha$, $\beta$, and $r$.  Due to space limitations, we report only the ACC and NMI results on the COIL20 dataset to evaluate parameter sensitivity. Fig.~\ref{ACCSensitivity} and Fig.~\ref{NMISensitivity} present the ACC and NMI performance of KAFUSE under varying values of $\alpha$, $\beta$, $r$ and feature selection ratios, respectively. We can observe that the parameters $\alpha$ and $\beta$ exhibit some sensitivity in our method with respect to both ACC and NMI. In contrast, the parameter $r$ is even more sensitive than $\alpha$ and $\beta$. The parameter $r$ needs to be tuned via grid search in practice to achieve optimal performance. Furthermore, the experimental results indicate that KAFUSE is relatively sensitive to the feature selection ratio. Determining the optimal feature selection ratio for each dataset remains a common and challenging problem.

\begin{table*}[t]
	\centering
	\caption{Performance comparison between KAFUSE and its two variants on eight datasets in terms of ACC and NMI.}\label{Table4}
	\resizebox{\textwidth}{!}{
	\renewcommand\tabcolsep{5 pt}
	\begin{tabular}{llcccccccc}
	\toprule
	\multirow{2}*{Metrics} & \multirow{2}*{Methods} & \multicolumn{8}{c}{Datasets} \\
	\cmidrule(r){3-10}
	~&~&NGs&Prokaryotic&COIL20&Cora&AD&CiteSeer&ALOI&Caltech\\
	\midrule
	\multirow{3}*{ACC}
	~& KAFUSE & $\mathbf{23.86\pm1.08} $ & $\mathbf{68.97\pm2.10} $ & $\mathbf{99.04\pm3.08} $ & $\mathbf{37.77\pm1.09} $ & $\mathbf{87.06\pm1.83} $ & $\mathbf{40.48\pm4.88} $ & $\mathbf{51.98\pm2.38} $ & $\mathbf{75.74\pm2.24} $\\
	~& KAFUSE-\uppercase\expandafter{\romannumeral 1}& $20.27\pm1.72 \bullet$ & $45.87\pm4.26 \bullet$ & $88.48\pm3.25 \bullet$ & $27.63\pm3.35 \bullet$ & $76.45\pm5.17 \bullet$ & $21.80\pm3.02 \bullet$ & $28.53\pm2.08 \bullet$ & $3.52\pm0.05 \bullet$ \\
	~& KAFUSE-\uppercase\expandafter{\romannumeral 2}& $21.64\pm1.05 \bullet$ & $35.72\pm6.32 \bullet$ & $92.46\pm3.48 \bullet$ & $23.55\pm1.45 \bullet$ & $70.97\pm5.49 \bullet$ & $23.91\pm2.36 \bullet$ & $31.40\pm2.12 \bullet$ &$3.50\pm0.04 \bullet$  \\
	\bottomrule
	
	\multirow{3}*{NMI}
	~& KAFUSE & $\mathbf{4.32\pm1.03} $ & $\mathbf{39.76\pm1.19} $ & $\mathbf{99.68\pm1.17} $ & $\mathbf{16.11\pm1.16} $ & $\mathbf{7.94\pm0.74} $ & $\mathbf{16.72\pm4.91} $ & $\mathbf{71.83\pm1.20} $ &$\mathbf{85.89\pm2.27} $ \\
	~& KAFUSE-\uppercase\expandafter{\romannumeral 1}& $1.56\pm1.35 \bullet$ & $6.40\pm1.23 \bullet$ & $95.79\pm1.82 \bullet$ & $1.85\pm1.55 \bullet$ & $2.39\pm0.62 \bullet$ & $0.88\pm2.79 \bullet$ & $66.23\pm1.48 \bullet$ & $6.84\pm0.51\bullet$  \\
	~& KAFUSE-\uppercase\expandafter{\romannumeral 2}& $1.71\pm1.21 \bullet$ & $24.92\pm1.75 \bullet$ & $97.32\pm2.59 \bullet$ & $10.48\pm1.89 \bullet$ & $0.92\pm0.23 \bullet$ & $2.35\pm1.68 \bullet$ & $64.54\pm1.85 \bullet$ & $6.70\pm0.48 \bullet$ \\
	\bottomrule
	
	\end{tabular}
	}
\end{table*}

\subsection{Visualization Analysis}
To evaluate the effectiveness of the selected features, t-SNE~\cite{van2008visualizing} is employed to project the samples onto a two-dimensional space for visualization. Fig.~\ref{tsne} presents the visualization results on the COIL20 dataset, where panel (a) shows the best-performing method (KAFUSE), and panels (b) and (c) display the second- and third-best methods, respectively. As shown, the visualization of our method reveals clusters that are both compact and well-separated when using the selected features. In contrast, the second- and third-best methods fail to clearly separate samples from the same class, resulting in these samples being intermixed and fragmented by samples from other classes. Therefore, these results demonstrate that our method can effectively preserve the local structure of the data with the selected features.

\subsection{Ablation Study}
In this subsection, we perform ablation experiments to assess the contribution of each component in the proposed KAFUSE. To this end, we introduce two variants of KAFUSE as follows:

1) KAFUSE-\uppercase\expandafter{\romannumeral 1}: It  utilizes only the cross-view consistent similarity graph learning component for multi-view feature selection.
\begin{equation}\label{5.5_1}	
	\begin{aligned}
		\min_{\tiny{\substack{\\\bm{W}^{(v)},\bm{\Lambda}^{(v)},\bm{F},\\\bm{Z},\bm{q}_{\cdot j},\bm{S}^{(v)},\bm{\theta}_v}}}&\sum_{v=1}^{V}\bm{\theta}_v^2\| {\bm{W}^{(v)}}^T \!\!\!\bm{\Lambda}^{(v)} \!\bm{X}^{(v)}\! \bm{H}\!\!-\!\!\bm{F} \bm{H} \|_F^2+\!\alpha\!\!\operatorname{Tr}(\!\bm{F}\!\bm{L}_Z\!\bm{F}^T\!)\\	
		&+\!\beta\operatorname{Tr}({\bm{\Lambda}}^{(v)} \!{\bm{X}}^{(v)} \!\bm{L}_{S^{(v)}} ({\bm{\Lambda}}^{(v)}\! {\bm{X}}^{(v)})^T\!)\!+\!{\gamma}^{(v)}\|{\bm{S}}^{(v)}\!\|_{F}^{2}\\
		&+\sum_{j=1}^{n}\|\bm{Z}_{. j}^{T}-\bm{q}_{\cdot j}^T \tilde{\bm{S}_j}\|_{2}^{2}+ \eta \|\bm{Z}\|_{F}^{2} 
	\end{aligned}
\end{equation}

2) KAFUSE-\uppercase\expandafter{\romannumeral 2}: It only uses the kernel alignment-based component for multi-view feature selection.
\begin{equation}\label{5.5_2}	
	\begin{aligned}
		\min_{\tiny{\substack{\bm{W}^{(v)},\bm{\Lambda}^{(v)},\\\bm{F},~\bm{\theta}_v,~\bm{\omega}_v}}} &\sum_{v=1}^{V}\bm{\theta}_v^2~\| {\bm{W}^{(v)}}^T \bm{\Lambda}^{(v)} \bm{X}^{(v)} \bm{H}-\bm{F} \bm{H} \|_F^2\\
        &-\bm{\omega}_v^r \operatorname{Tr}(\bm{H}\bm{K}_c^{(v)} \bm{H} \bm{K}_{u}^{(v)})
	\end{aligned}
\end{equation}

For both KAFUSE-\uppercase\expandafter{\romannumeral 1} and KAFUSE-\uppercase\expandafter{\romannumeral 2}, the constraints imposed on the corresponding variables are the same as those specified in Eq.~(\ref{4_2}). Table~\ref{Table4} presents a comparison of the performance of KAFUSE and its two variants on eight datasets with respect to ACC and NMI. It can be seen from Table~\ref{Table4} that KAFUSE achieves the best performance in both ACC and NMI. Compared to KAFUSE-\uppercase\expandafter{\romannumeral 1}, KAFUSE achieves significantly better performance. This demonstrates the effectiveness of the kernel alignment module in reducing feature redundancy and enabling the removal of redundant features. Additionally, KAFUSE demonstrates superior performance compared to KAFUSE-\uppercase\expandafter{\romannumeral 2} across all datasets, indicating that corss-view consistent similarity graph learning with auto-weighted sample-level fusion is beneficial for improving feature selection performance.

\section{Conclusion}
In this paper, we proposed a novel multi-view unsupervised feature selection method called KAFUSE, which utilizes kernel alignment-based redundancy reduction and cross-view consistent graph learning with sample-level weights. The proposed KAFUSE reduce feature redundancy in both linear and nonlinear ways by utilization of the orthogonal constraint and kernel alignment, thereby selecting the uncorrelated and discriminative features. Meanwhile, a similarity graph is adaptively learned for each view in the selected feature space. The resulting view-specific graphs are stacked into a tensor and fused via sample-level automatic weighting, facilitating the learning of a consistent similarity graph across different views and effectively preserving the data's local structure. Extensive experimental results on eight real-world multi-view datasets demonstrate the efficacy of KAFUSE compared to state-of-the-art methods.

\bibliographystyle{IEEEtran}
\bibliography{refabb}

\end{document}